\documentclass[10pt,twocolumn,letterpaper]{article}
\usepackage[pagenumbers]{cvpr}
\usepackage{cuted} %

\usepackage{tabularx}
\usepackage{multirow}
\usepackage[utf8]{inputenc}
\usepackage[T1]{fontenc}
\usepackage{amsfonts}
\usepackage{nicefrac}
\usepackage{microtype}
\usepackage{colortbl}
\usepackage{fontawesome5}
\definecolor{cvprblue}{rgb}{0.21,0.49,0.74}
\usepackage[hyperfootnotes=false,breaklinks,colorlinks,allcolors=cvprblue]{hyperref}

\title{Aes3D: Aesthetic Assessment in 3D Gaussian Splatting}

\author{%
  \textbf{Chuanzhi Xu}\textsuperscript{1,*\textdagger}\quad
  \textbf{Boyu Wei}\textsuperscript{1,*}\quad
  \textbf{Haoxian Zhou}\textsuperscript{1}\quad
  \textbf{Xuanhua Yin}\textsuperscript{1} \\
  \textbf{Zihan Deng\textsuperscript{2}}\quad
  \textbf{Haodong Chen\textsuperscript{1}}\quad
  \textbf{Qiang Qu\textsuperscript{1}}\quad
  \textbf{Weidong Cai\textsuperscript{1}} \\
  \textsuperscript{1}The University of Sydney\quad
  \textsuperscript{2}The University of Hong Kong\quad
  \href{https://github.com/LouckXu/Aes3D}{\faGithub\ Project Page}
}

\hypersetup{
  pdftitle={Aes3D: Aesthetic Assessment in 3D Gaussian Splatting},
  pdfauthor={Chuanzhi Xu, Boyu Wei, Haoxian Zhou, Xuanhua Yin, Zihan Deng, Haodong Chen, Qiang Qu, Weidong Cai},
  pdfsubject={3D scene aesthetic assessment},
  pdfkeywords={3D Aesthetic Assessment, 3D Gaussian Splatting, Image Aesthetic Assessment, 3D Scene Understanding, Multi-view Learning}
}

\begin{document}

\maketitle
{
\renewcommand{\thefootnote}{}
\makeatletter
\renewcommand{\@makefntext}[1]{\noindent#1}
\makeatother
\footnotetext{\raggedright\textsuperscript{*}Equal contribution.\\\textsuperscript{\textdagger}Corresponding author: \texttt{chuanzhi.xu@sydney.edu.au}}
}
\begin{strip}
    \vspace{-50pt}
    \centering
    \includegraphics[width=\linewidth]{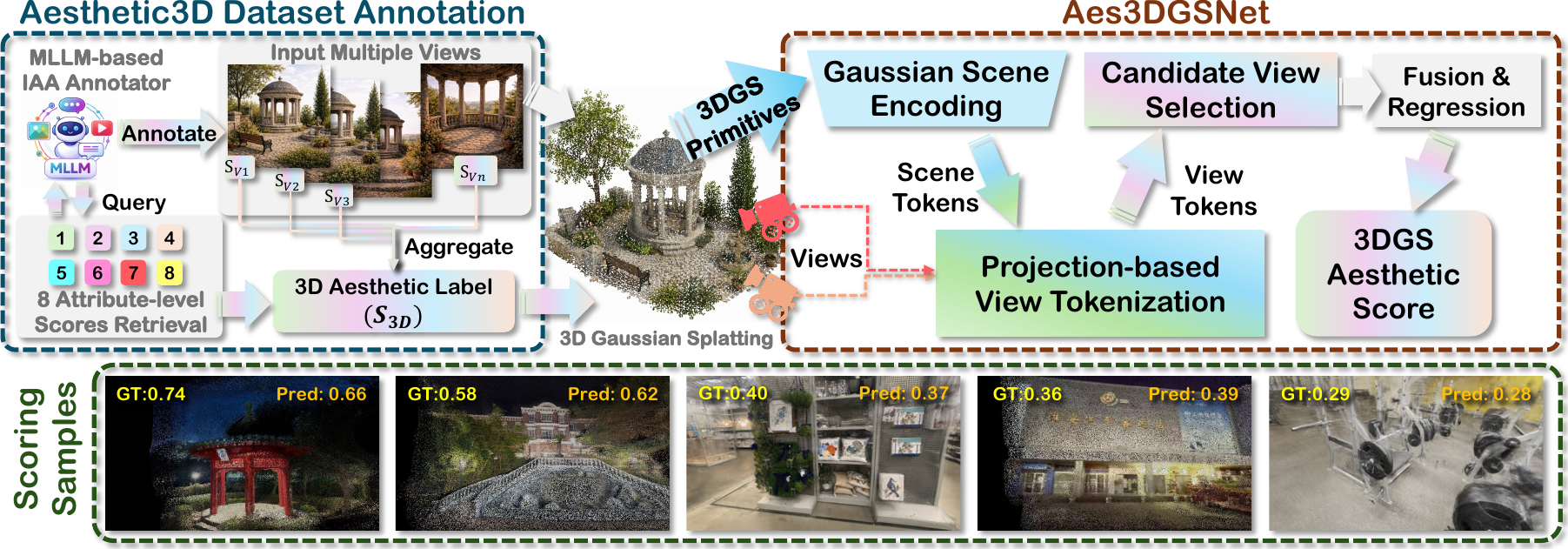}
    \captionsetup{hypcap=false}
    \captionof{figure}{Overview of \textbf{Aes3D}: \textbf{Aesthetic3D} is built through IAA-based annotation of multiple rendered views, while \textbf{Aes3DGSNet} predicts scene-level aesthetic scores from 3D Gaussian primitives. Bottom: examples compare ground-truth (GT) and predicted (Pred) scores.}
    \captionsetup{hypcap=true}
    \label{fig:teaser}
\end{strip}
\begin{abstract}
As 3D Gaussian Splatting (3DGS) gains attention in immersive media and digital content creation, assessing the aesthetics of 3D scenes becomes important in helping creators build more visually compelling 3D content. However, existing evaluation methods for 3D scenes primarily emphasize reconstruction fidelity and perceptual realism, largely overlooking higher-level aesthetic attributes such as composition, harmony, and visual appeal. This limitation comes from two key challenges: (1) the absence of general 3DGS datasets with aesthetic annotations, and (2) the intrinsic nature of 3DGS as a low-level primitive representation, which makes it difficult to capture high-level aesthetic features.
To address these challenges, we propose \textbf{Aes3D}, the first systematic framework for assessing the aesthetics of 3D neural rendering scenes. Aes3D includes \textbf{Aesthetic3D}, the first dataset dedicated to 3D scene aesthetic assessment, built on our proposed annotation strategy for 3D scene aesthetics. In addition, we present \textbf{Aes3DGSNet}, a lightweight model that directly predicts scene-level aesthetic scores from 3DGS representations. Notably, our model operates solely on 3D Gaussian primitives, eliminating the need for rendering multi-view images and thus reducing computational cost and hardware requirements.
Through aesthetics-supervised learning on multi-view 3DGS scene representations, Aes3DGSNet effectively captures high-level aesthetic cues and accurately regresses aesthetic scores. Experimental results demonstrate that our approach achieves strong performance while maintaining a lightweight design, establishing a new benchmark for 3D scene aesthetic assessment.

\end{abstract}

\section{Introduction}

The recent development of 3D scene neural rendering and novel view synthesis (NVS), particularly 3D Gaussian Splatting (3DGS), has opened new opportunities for visual perception research \cite{GS00, GS00-1}. 3DGS explicitly models scenes using a set of anisotropic 3D Gaussian distributions, enabling high-quality, real-time novel view synthesis \cite{GS01}. Compared to Neural Radiance Fields (NeRF) \cite{GS02}, 3DGS significantly reduces training and rendering costs while maintaining or even improving rendering quality, demonstrating strong potential for applications in virtual reality, digital twins, and immersive media \cite{GS00, GS00-1}. As a result, assessing the perceptual quality of 3DGS-generated content has emerged as a research direction. Existing evaluation methods for 3DGS have mainly focused on reconstruction accuracy and perceptual quality metrics \cite{GSP01,GSP02}. Traditional metrics such as PSNR, SSIM, and LPIPS are widely used to measure reconstruction performance \cite{GS01}, but these metrics often fail to align with human perception in complex scenes. Recent studies have begun introducing subjective evaluation datasets and perceptual metrics \cite{GSP01, GSP02, GSP03, GSP04, GSP05} to better assess human visual experience, but most of these works still focus on the level of "realism" or "distortion degree" and do not reach higher-level visual assessment.

Image Aesthetic Assessment (IAA) has long been an important problem in computer vision, aiming to quantify subjective quality in human visual perception beyond low-level fidelity \cite{IAA1, IAA2}. Early work focused on 2D image aesthetics, capturing factors such as composition, color harmony, and lighting \cite{IAA3, IAA4, IAA5}. With advances in deep learning, large-scale datasets such as AVA and AADB \cite{IAA6, ASS2} have driven data-driven approaches, enabling CNN and Transformer-based models to learn aesthetic mappings from visual features \cite{IAA8, IAA7}. Video aesthetic assessment further introduces temporal complexity, including temporal consistency, motion dynamics, and shot transitions \cite{IAA9, IAA10}. In parallel, aesthetics-driven image enhancement and color grading have emerged, where models assess and optimize visual appearance according to human preferences \cite{IAE01, IAE02, IAE03, yin2026accelaes}.

However, aesthetic quality is not equivalent to perceptual quality. Aesthetic quality focuses on high-level, holistic semantic attributes such as visual appeal, artistic expressiveness, composition, color, and emotional expression \cite{IAA12, IAARE08}. This distinction is particularly important in 3D scenes: a 3D scene reconstruction may achieve high accuracy in geometry and texture, but still lacks visual appeal or artistic expressiveness. Although some works have begun to explore 3D optimization based on human preferences \cite{GSE01, GSE02} or stylization methods \cite{GST01, GST02}, there is no systematic framework for the aesthetic assessment of 3D neural-rendered scenes.

We believe that 3D aesthetic assessment is inherently challenging. First, no general 3D scenes dataset with aesthetic-quality labels currently exists to support explicit modeling of aesthetic scores. Second, compared to 2D evaluation, the aesthetic quality of a 3D scene should emerge from the combination of aesthetic perceptions across multiple viewpoints, yet these perceptions may vary significantly across views. Furthermore, in the context of 3DGS aesthetic assessment, aesthetic quality depends not only on the rendered results but also on the underlying properties of the Gaussian distribution, including density, color, and geometry. We anticipate that addressing these challenges will be significant for enabling aesthetics-driven applications such as 3D rendering, editing, compression, and generation, and may provide aesthetic guidance for 3D animation and game graphics.

In this paper, we propose Aes3D, the first academic study on assessing aesthetics for 3D neural-rendered scenes across general scene categories, including the Aesthetic3D dataset and the Aes3DGSNet, as shown in Figure \ref{fig:teaser}. Our contributions can be summarized as:
\begin{itemize}
    \item \textbf{Aesthetic3D Dataset.} We construct the first 3D scene aesthetic dataset, Aesthetic3D, by designing an IAA-based annotation pipeline that employs a multimodal large language model (MLLM)-based aesthetic annotator to produce both overall aesthetic scores and fine-grained attribute-level sub-scores across 8 aesthetic dimensions, supporting research on 3D aesthetic assessment.
    \item \textbf{Aes3DGSNet.} We propose Aes3DGSNet, a novel network for scene-level aesthetic assessment from input of 3DGS primitives, enabling rendering-independent and view-aware aesthetic score prediction. It achieves general 3D aesthetic assessment for the first time.
    \item \textbf{Benchmark and Evaluation.} Aes3D establishes the first benchmark for 3DGS aesthetic assessment based on Aesthetic3D, and shows that Aes3DGSNet achieves superior aesthetic assessment performance with a lightweight architecture, demonstrating the effectiveness and research significance of Aes3D.

\end{itemize}

\begin{figure*}[t]
    \centering
    \includegraphics[width=\textwidth]{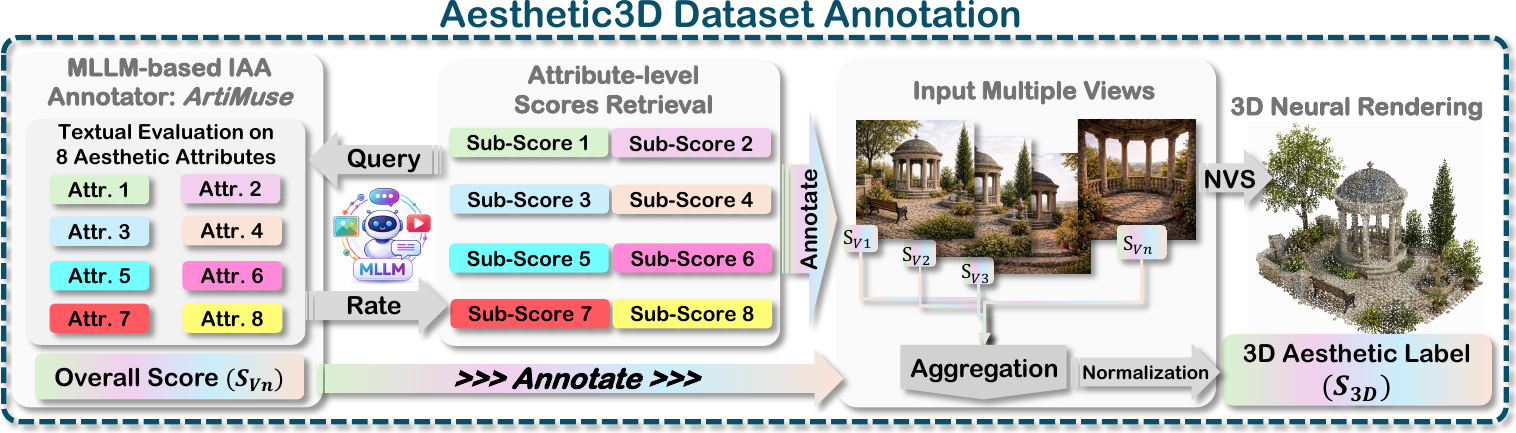}
    \caption{IAA-based annotation pipeline for constructing Aesthetic3D. ArtiMuse assigns overall and eight attribute-level aesthetic scores to multiple rendered views. The view-level scores are then aggregated and normalized into a scene-level label.}
    \label{fig:anno}
\end{figure*}

\section{Related Work}
\noindent\textbf{Image Aesthetic Assessment.}
Early studies on image aesthetic assessment rely on handcrafted features derived from photographic rules such as composition, color harmony, and depth of field \cite{IAA3,IAA4,IAA5}. With the advent of deep learning, CNN-based methods enable automatic aesthetic feature learning, with early works \cite{IAARE02,IAARE03} validating their effectiveness and the large-scale AVA dataset \cite{IAA6} further driving progress. NIMA \cite{IAA11} improves subjectivity modeling by predicting score distributions instead of scalar scores. Subsequent research focuses on enhancing feature representation, including multi-patch aggregation and multi-level spatial pooling \cite{IAARE06,IAARE07}, as well as attention mechanisms that emphasize salient regions \cite{IAARE08}. More recently, advanced architectures and paradigms have been explored, such as graph-based modeling of region relationships \cite{IAARE09}, Transformer-based methods like MUSIQ \cite{IAARE10} for long-range dependency modeling, and multimodal approaches (e.g., VILA \cite{IAARE11}) that incorporate textual cues. Additionally, several works \cite{IAARE12,IAARE13,IAARE14} consider user preference modeling for personalized assessment. Comprehensive surveys \cite{IAA1,IAA2,IAA1-1} further highlight the significance and practical value of image aesthetic assessment. However, all of these methods rely on single-view static images, whereas 3D representations, such as NeRF or 3DGS, exhibit multi-view consistency, spatial structure, and viewpoint-dependent appearance changes. Thus, 3D aesthetic assessment considers individual view quality, overall perception across views, and spatial coherence.

\noindent\textbf{Perceptual Quality Assessment of 3D Neural Rendering (NeRF \& 3DGS).}
With the rapid development of neural rendering methods (or NVS) such as NeRF and 3DGS, there is an increasing demand for reliable perceptual quality assessment. Early studies mainly relied on full-reference image quality metrics computed on rendered novel views, such as PSNR, SSIM, and LPIPS~\cite{GS01,GS02}. However, these metrics often show weak correlation with human perception in complex scenarios involving view-dependent effects and temporal inconsistencies. For NeRF, many works constructed subjective datasets and employed pairwise comparison protocols to systematically analyze the perceptual quality of different NVS methods and their correlation with objective metrics, revealing that traditional metrics fail to reliably reflect video-level perceptual quality~\cite{N3,N5,N4}. More recent approaches propose quality modeling frameworks that combine view-level and point-level features, enabling no-reference perceptual quality prediction without relying on ground-truth images~\cite{N1,N6}. For 3DGS, mainstream methods also build dedicated datasets and design sampling strategies to render multi-view images or videos, in order to analyze the impact of different distortion factors (e.g., compression, resolution, and reconstruction errors) on visual quality~\cite{GSP04,GSP05,GSP01,baseline4,GSP03}. More recent work further constructs synthetic datasets to explore no-reference quality prediction directly on 3D Gaussian primitives, avoiding the need for rendered images~\cite{GSP02}. However, these evaluation paradigms rely heavily on viewpoint sampling strategies and camera trajectory design, making the evaluation more about the quality of the viewing path than the intrinsic quality of the 3D scene itself. Moreover, existing studies mainly focus on perceptual fidelity, such as structural consistency, texture clarity, and temporal stability, while paying limited attention to higher-level visual attributes, including aesthetic quality, composition, and visual appeal.

\section{Aesthetic3D Dataset}
\label{sec:annotation}
\begin{figure*}[t]
    \centering
    \includegraphics[width=\textwidth]{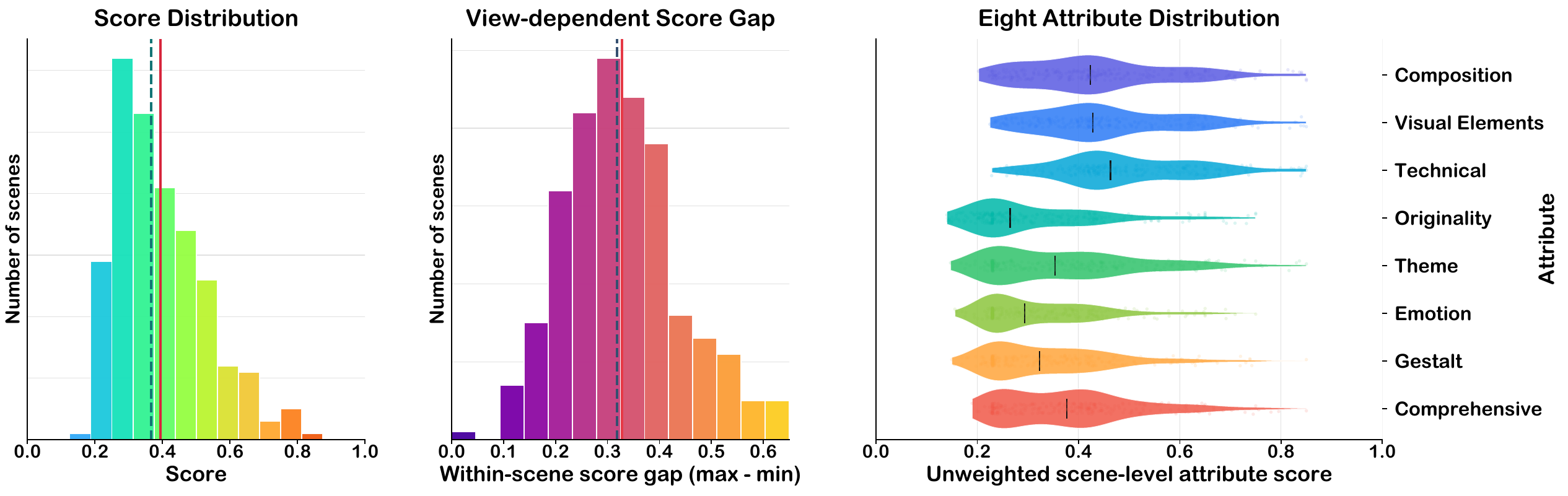}
    \caption{Statistical overview of Aesthetic3D (8-attr mean). \textbf{Left:} distribution of scene-level scores. \textbf{Middle:} distribution of within-scene view-dependent score gaps, computed as the difference between the maximum and minimum view-level scores. \textbf{Right:} distributions of the eight unweighted scene-level attribute scores. Dashed lines indicate medians. Red lines indicate means.}
    \label{fig:statistic}
\end{figure*}
\subsection{Motivation \& Overview}

One principled approach to aesthetic assessment of 3D scenes is to construct a dedicated dataset with human-annotated aesthetic scores. However, collecting such a dataset is highly expensive and labor-intensive, as it requires large-scale user studies across diverse scenes and viewpoints. On the other hand, aesthetic assessment for 2D images has been extensively studied, with mature models and well-established datasets. We believe that human aesthetic perception of 3D scenes is largely formed by aggregating aesthetic impressions from multiple viewpoints, combined with an additional sense of spatial structure unique to 3D. Importantly, the aesthetic perception at each individual viewpoint can still be reasonably approximated as a 2D image assessment problem. This is consistent with prior work in neural rendering perceptual quality assessment \cite{N3,N5,N4,GSP04,GSP05,GSP01}, where 3D scene quality is commonly evaluated via rendered image or video sequences from multiple viewpoints.

Motivated by this observation, we approximate 3D scene aesthetics by leveraging a state-of-the-art 2D aesthetic assessment model, ArtiMuse \cite{ASS3}, as the annotator to score the original source images of 3D scene datasets. These scores are treated as proxy labels of human aesthetic perception at each viewpoint. We then aggregate the scores across all viewpoints to obtain a scene-level aesthetic score for each 3D rendering scene. Following this pipeline, we annotate two 3D scene datasets (DL3DV-10K \cite{DT2}, and Bilarf \cite{DT3}), merge the scored samples, and construct a new benchmark dataset, termed the Aesthetic3D. An annotation schematic diagram can be found in Figure \ref{fig:anno}.

\noindent\textbf{Introduction of Selected IAA Annotator.}
ArtiMuse \cite{ASS3} is an MLLM-based image aesthetic assessment framework that jointly performs fine-grained aesthetic scoring and expert-level textual analysis. It is trained on ArtiMuse-10K, a dataset of 10,000 expert-annotated images evaluated across eight aesthetic dimensions: \textit{Composition \& Design, Visual Elements \& Structure, Technical Execution, Originality \& Creativity, Theme \& Communication, Emotion \& Viewer Response, Overall Gestalt, and Comprehensive Evaluation}. To enable precise fine-grained score prediction, ArtiMuse introduces Token As Score, a strategy that maps 101 ordered tokens from the existing tokenizer vocabulary to integer scores from 0 to 100, avoiding vocabulary expansion and quantization loss. Trained in a two-stage pipeline of text pretraining followed by LoRA-based score fine-tuning, ArtiMuse achieves state-of-the-art performance on multiple IAA benchmarks including AVA \cite{IAA6}, PARA \cite{IAARE12}, TAD66K \cite{ASS1}, and FLICKR-AES \cite{ASS1-DT2}. Furthermore, ArtiMuse demonstrates strong generalization capabilities across various image types (such as photography, paintings, and AIGC-generated images), and its multi-dimensional attribute modeling approach better aligns with the inherently multi-factorial nature of human aesthetics. For detailed explanation, please refer to the supplementary material.

\noindent\textbf{Introduction of Selected 3D Scene Datasets.}
To construct the Aesthetic3D dataset, we select two widely used
3D scene datasets, DL3DV-10K (sampled subset) and Bilarf. By covering diverse scene types and visual
characteristics, ranging from large-scale real-world environments to scenes with challenging
lighting conditions, the combined datasets enable robust evaluation of aesthetic perception in 3DGS and
improve the generalizability of learned models. More importantly, they reflect the types of 3D
scenes that an individual user or a game/film producer would likely provide when seeking to
evaluate scene aesthetics. For detailed explanation, please refer to the supplementary material.

\subsection{Aesthetic Annotation \& Score Aggregation}

\noindent\textbf{Aesthetic Annotation \& Attribute-level Scores Retrieval.} ArtiMuse provides a global aesthetic score together with textual evaluations over 8 aesthetic attributes. To obtain quantitative attribute-level labels, we further leverage its conversational inference interface and perform 8 independent attribute-scoring queries for each input view. For each attribute, we design a unified prompt: “Rate the aesthetic quality of this image from the aspect of $\{attr\_name\}$ on a 0–100 scale. Output only one number.” to query a single aesthetic score from the corresponding attribute perspective. The same image is then paired with 8 attribute prompts, producing 8 scoring outputs. This design has several advantages. First, it avoids training additional regression heads and directly uses the fine-grained aesthetic understanding already learned by ArtiMuse, effectively turning a generative model into a structured attribute scorer. Second, each attribute is evaluated using an explicit prompt, which improves interpretability and ensures that each score has a clear semantic meaning. Third, the 8 attribute scores are obtained through independent queries rather than a single coupled forward pass, making each score closer to an isolated judgment of one aesthetic factor, which is more suitable for attribute-level supervision. In the data analysis and the supplementary material, we show that the average of the 8 attribute scores (8-attr mean) is highly consistent with the global score provided by ArtiMuse, which supports the reliability of this design. In summary, each input view will be annotated with an overall score as well as sub-scores across 8 attributes. Meanwhile, we also retain the textual evaluations as view-level labels for potential future use in aggregating subjective aesthetic descriptions of 3D scenes.

\noindent\textbf{Aesthetic Scores Aggregation.} At the scene level, we use the overall score or 8-attr mean score as the main label. Specifically, we average the global scores across all input views of a 3D scene to obtain the final scene-level aesthetic score, and then normalize it to the $[0,1]$ range.

\subsection{Analysis of Data Statistics}

To validate the reliability of the constructed supervision, we analyze the statistical properties of the Aesthetic3D dataset. Please find Figure \ref{fig:statistic} for an overview and the supplementary material for more detailed statistics. Aesthetic3D contains 278 reconstructed scenes and 92,649 input views, averaging 333 views per scene. The scene-level aesthetic score (8-attr mean) exhibits a non-degenerate distribution, mainly spanning from around 0.180 to 0.825 with a mean of 0.394, a median of 0.367, and a standard deviation of 0.133. This indicates that the dataset covers a meaningful range of aesthetic quality instead of collapsing to a narrow score interval. A more important observation is the strong view dependency of 3D scene aesthetics. For each scene, we measure the within-scene score gap as the difference between the maximum and minimum view-level scores. The resulting gap has a mean 0.33, a median 0.32, a 90th percentile 0.47, and a maximum 0.85. Moreover, 83.2\% of scenes have a score gap larger than 0.20, and 49.3\% exceed 0.30. These statistics confirm that aesthetic perception varies substantially across viewpoints, motivating a multi-view assessment paradigm rather than single-view prediction. We further examine the 8 attribute-level labels. The eight attributes show distinct marginal distributions: \textit{technical execution} has the highest mean score but the lowest discriminative ratio, whereas \textit{theme}, \textit{emotion}, \textit{gestalt}, and \textit{comprehensive evaluation} exhibit stronger discrimination across scenes. At the same time, the attributes remain highly correlated, with inter-attribute Spearman correlations ranging from 0.933 to 0.994, and their unweighted mean is highly consistent with the holistic aesthetic score ($r{=}0.992$, $\rho{=}0.993$). This suggests that the attribute scores should be understood as an interpretable decomposition of a shared aesthetic signal, rather than as independent supervision targets. In addition, we also conducted a human study to validate the proxy aesthetic labels. Please refer to the supplementary material.

\section{Aes3DGSNet}
\label{sec:method}

We propose Aes3DGSNet to address scene-level aesthetic assessment for 3D Gaussian Splatting. Given a reconstructed scene represented by a set of
Gaussian primitives $\mathcal{G}$, the goal is to predict a scalar aesthetic
score $\hat{y}$. While 3DGS provides an explicit representation of scene
geometry and appearance, many aesthetic cues, such as composition, occlusion,
spatial arrangement, object distribution, and visual balance, are inherently
view-dependent. A single global 3D summary may miss view-conditioned
evidence, while averaging all candidate views may dilute useful observations
with redundant or low-quality views.

In practice, the input multiple views used during Aesthetic3D annotation may not
coincide with those available at inference time, and rendered images from novel
viewpoints can suffer from blur, artifacts, and geometric distortions that
severely compromise aesthetic perception. Therefore, instead of taking rendered
RGB images as direct model inputs, Aes3DGSNet performs aesthetic assessment
directly on Gaussian primitives and derives view-level evidence from view geometry through
geometric projection without rendering images. This design reduces dependence on
viewpoint-specific rendering artifacts and encourages the model to rely on the
intrinsic 3D scene representation.

\begin{figure*}[t]
    \centering
    \includegraphics[width=\textwidth]{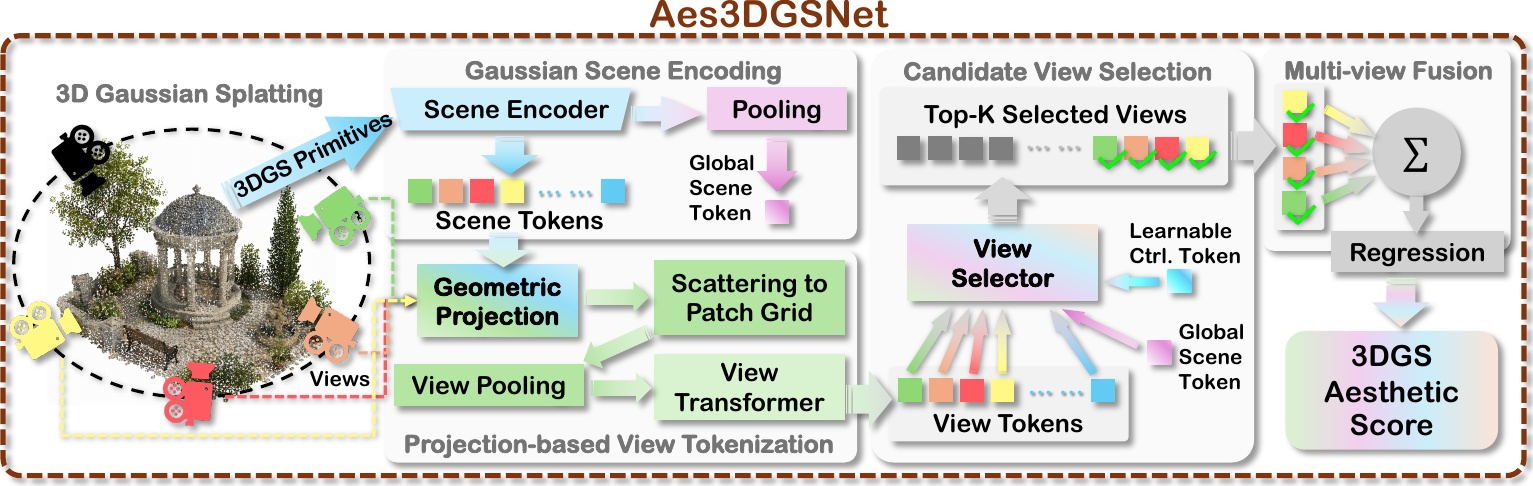}
    \caption{Overview of Aes3DGSNet. Gaussian primitives are encoded into scene tokens and projected onto candidate-view patch grids. A learned top-$K$ selector retains informative views, whose features are fused to predict the scene-level aesthetic score.}
    \label{fig:net}
\end{figure*}

Therefore, we formulate the 3DGS aesthetic assessment as a view-selective evidence aggregation problem. The model first tokenizes Gaussian primitives into 3D scene tokens (S\ref{4.1}), then projects these tokens into candidate views to construct view descriptors (S\ref{4.2}), and finally learns to select a compact subset of informative views (S\ref{4.3}) for multi-view fusion and scene-level regression (S\ref{4.4}). Figure \ref{fig:net} provides an overview of Aes3DGSNet.

\subsection{Gaussian Scene Encoding}
\label{4.1}
We represent each 3DGS scene using a fixed-size subset of Gaussian primitives. In our default setting, we use an FPS-subsampled ~\citep{qi2017pointnet++} set of $N$ Gaussians for efficient and spatially balanced scene encoding. For Gaussian $i$, the input is
$\mathbf{x}_i=[\mathbf{p}_i,\mathbf{c}_i,\bar{\mathbf{p}}_i]$, where
$\mathbf{p}_i \in \mathbb{R}^{3}$ denotes the normalized Gaussian center,
$\mathbf{c}_i \in \mathbb{R}^{3}$ denotes its RGB attribute, and
$\bar{\mathbf{p}}_i=\mathbf{p}_i/\|\mathbf{p}_i\|_2$ is the unit direction
from the scene center.

The input feature is mapped to a hidden token and augmented with a
coordinate-based positional embedding as:
\begin{equation}
\mathbf{h}_i^{0}=\phi_{\mathrm{in}}(\mathbf{x}_i)+\phi_{\mathrm{pos}}(\mathbf{p}_i).
\end{equation}
Then, the scene encoder, a point Transformer encoder, models contextual interactions among Gaussian
tokens and outputs point-level scene tokens:
\begin{equation}
\mathbf{T}=\mathrm{Enc}_{3D}(\{\mathbf{h}_i^{0}\}_{i=1}^{N})
\in\mathbb{R}^{N\times D}.
\end{equation}
Attention pooling also produces a global scene token $\mathbf{s}$ for view selection.

The scene encoder follows the architectural philosophy of Point Transformer V3
and related point representation learning methods~\citep{wu2024point,wu2025sonata},
but all model parameters are trained from scratch without using pretrained
point-cloud, image, or multimodal weights.

\subsection{Projection-based View Tokenization}
\label{4.2}
To obtain view-level evidence without rendering RGB images, we project the 3D scene tokens into candidate views. This design is motivated by 3D-to-multi-view representation learning~\citep{chen2025point}, where camera geometry is used to derive multi-view evidence from 3D point tokens. For the $v$-th candidate view, each 3D point is projected onto the view plane
using the corresponding intrinsic matrix $\mathbf{K}_v$, expressed in
normalized image coordinates, and world-to-camera extrinsic matrix
$\mathbf{E}_v=[\mathbf{R}_v \mid \mathbf{t}_v]$:
\begin{equation}
    \mathbf{u}_{iv}
    =
    \mathbf{K}_v
    \Pi
    \left(
        \mathbf{E}_v
        \tilde{\mathbf{p}}_i
    \right),
\end{equation}
where $\tilde{\mathbf{p}}_i=[\mathbf{p}_i^\top,1]^\top$ is the homogeneous form
of $\mathbf{p}_i$, and $\Pi([x,y,z]^\top)=[x/z,\;y/z,\;1]^\top$ denotes
perspective division to normalized image coordinates. The projected view plane is discretized into a $G \times G$ patch grid. Let
$\mathcal{T}_{v,m}$ be the set of scene tokens projected into patch $m$ of
view $v$. For each patch, we first aggregate the projected tokens into a pooled
representation $\mathbf{b}_{v,m}$ using mean and max pooling (View Pooling),
and use a learned empty-cell token when $\mathcal{T}_{v,m}=\emptyset$. We then
construct the geometry-aware patch token as:
\begin{equation}
    \mathbf{r}_{v,m}
    =
    \phi_{\mathrm{cell}}(\mathbf{b}_{v,m})
    +
    \mathbf{a}_v,
\end{equation}
where $\mathbf{a}_v$ is a view-conditioned context vector derived from the
camera extrinsics and normalized intrinsics of view $v$. The resulting patch tokens are refined by a lightweight view Transformer and
summarized into a single view token descriptor $\mathbf{v}_v$. Thus, each
descriptor is constructed from projected 3D scene tokens conditioned on view
geometry, rather than from rendered image features or a pretrained 2D backbone.
See the supplementary material for full details.

\subsection{Candidate View Selection}
\label{4.3}
Different candidate views contribute unequally to 3D-scene-level aesthetic
assessment. Some views expose composition, spatial structure, and object
relationships more clearly, while others may be redundant, occluded, or weakly
informative. Therefore, we introduce a learned top-$K$ selector to aggregate a limited set of informative view descriptors instead of uniformly averaging all
views.

The view selector is inspired by Flexible Frame Selection (FFS)~\citep{buch2025flexible}, which learns to select
informative frames for efficient video reasoning. We
adapt this idea from temporal frame selection to candidate view selection in
3D scenes. Each view descriptor is first mapped to the selector space as
$\mathbf{z}_v=\phi_{\mathrm{sel}}(\mathbf{v}_v)$, forming the candidate view tokens fed into the selector. The selector then forms a
sequence containing the global scene token, learnable control tokens (inspired by the [CTRL] register in FFS), and all
candidate view tokens,
$\mathbf{Z}=[\phi_{\mathrm{s}}(\mathbf{s}),\mathbf{c}_1,\ldots,\mathbf{c}_M,
\mathbf{z}_1,\ldots,\mathbf{z}_V]$.
A Transformer performs listwise refinement over this sequence, and a utility
head predicts the importance score $u_v=\psi(\mathbf{z}'_v)$ for each refined
view token. To constrain the number of selected views, we compute sparse top-$K$ selection weights over valid
candidate views:
\begin{equation}
    \alpha_v
    =
    \frac{
        \exp(u_v/\tau)\,
        \mathbb{1}[v\in\mathrm{TopK}(u)]
    }{
        \sum_{k\in\mathcal{V}}
        \exp(u_k/\tau)\,
        \mathbb{1}[k\in\mathrm{TopK}(u)]
    },
\end{equation}
where $\mathcal{V}$ denotes the set of valid views and $\tau$ is a temperature
parameter. This mechanism constrains the regression path to rely on a compact
subset of views while preserving trainability through the soft weighting
branch.

\definecolor{blockA}{RGB}{255, 245, 235}
\definecolor{blockB}{RGB}{255, 250, 230}
\definecolor{blockC}{RGB}{240, 250, 240}
\definecolor{blockD}{RGB}{235, 245, 255}

\begin{table*}[t]
\centering
\caption{
Main results on scene aesthetic regression. Best results in each column are in \textbf{bold}.
}
\label{tab:main_results}
\setlength{\tabcolsep}{1.5pt}
\resizebox{0.72\textwidth}{!}{%
\begin{tabular}{l| r r | c c c c c}
\toprule
Method & \#Params & GFLOPs & PLCC $\uparrow$ & SRCC $\uparrow$ & KRCC $\uparrow$ & MAE $\downarrow$ & RMSE $\downarrow$ \\
\midrule

\rowcolor{blockA}
\multicolumn{8}{l}{\textit{\textbf{(A) Zero-shot 2D Image Aesthetic Baselines}}}\\
\rowcolor{blockA}
ArtiMuse
& $\sim\!8$B & $\sim257.7$K
& $.224\mathord{\pm}.094$
& $.207\mathord{\pm}.072$
& $.143\mathord{\pm}.051$
& $.106\mathord{\pm}.006$
& $.125\mathord{\pm}.007$ \\
\rowcolor{blockA}
AADB
& $70.9$M & $2.4$
& $-.020\mathord{\pm}.124$
& $-.032\mathord{\pm}.103$
& $-.021\mathord{\pm}.076$
& $.111\mathord{\pm}.007$
& $.134\mathord{\pm}.007$ \\
\rowcolor{blockA}
NIMA
& $14.7$M & $15.4$
& $.081\mathord{\pm}.059$
& $-.038\mathord{\pm}.113$
& $-.028\mathord{\pm}.078$
& $.143\mathord{\pm}.010$
& $.167\mathord{\pm}.010$ \\
\rowcolor{blockA}
TANet
& $13.9$M & $4.1$
& $.075\mathord{\pm}.149$
& $-.015\mathord{\pm}.176$
& $-.010\mathord{\pm}.118$
& $.116\mathord{\pm}.008$
& $.136\mathord{\pm}.007$ \\

\addlinespace[2pt]

\rowcolor{blockB}
\multicolumn{8}{l}{\textit{\textbf{(B) 3DGS Quality Assessment Baselines}}}\\
\rowcolor{blockB}
MUGSQA-DBCNN
& $15.3$M & $712.2$
& $.311\mathord{\pm}.089$
& $.343\mathord{\pm}.069$
& $.233\mathord{\pm}.047$
& $.105\mathord{\pm}.005$
& $.130\mathord{\pm}.004$ \\
\rowcolor{blockB}
GSOQA
& $27.5$M & $3.5$
& $.503\mathord{\pm}.044$
& $.485\mathord{\pm}.067$
& $.351\mathord{\pm}.046$
& $.092\mathord{\pm}.003$
& $.119\mathord{\pm}.005$ \\

\addlinespace[2pt]

\rowcolor{blockC}
\multicolumn{8}{l}{\textit{\textbf{(C) Point-based Quality Assessment Baselines}}}\\
\rowcolor{blockC}
MM-PCQA
& $52.9$M & $79.9$
& $.441\mathord{\pm}.034$
& $.420\mathord{\pm}.060$
& $.286\mathord{\pm}.048$
& $.093\mathord{\pm}.002$
& $.117\mathord{\pm}.002$ \\
\rowcolor{blockC}
LRL-GQA
& $17.4$K & $8.1$
& $.409\mathord{\pm}.118$
& $.378\mathord{\pm}.076$
& $.255\mathord{\pm}.056$
& $.097\mathord{\pm}.008$
& $.118\mathord{\pm}.006$ \\
\rowcolor{blockC}
Stochastic NR-PCQA
& $25.6$M & $6123.7$
& $.509\mathord{\pm}.011$
& $.488\mathord{\pm}.033$
& $.335\mathord{\pm}.014$
& $.090\mathord{\pm}.001$
& $.112\mathord{\pm}.001$ \\

\midrule
\rowcolor{blockD}
\textbf{Aes3DGSNet (total)}
& $3.2$M & $29.4$
& $.540\mathord{\pm}.028$
& $.518\mathord{\pm}.049$
& $.362\mathord{\pm}.042$
& $\mathbf{.076\mathord{\pm}.012}$
& $\mathbf{.101\mathord{\pm}.013}$ \\
\rowcolor{blockD}
\textbf{Aes3DGSNet (8-attr)}
& $3.2$M & $29.4$
& $\mathbf{.567\mathord{\pm}.036}$
& $\mathbf{.554\mathord{\pm}.043}$
& $\mathbf{.389\mathord{\pm}.032}$
& $.084\mathord{\pm}.008$
& $.110\mathord{\pm}.005$ \\

\bottomrule
\end{tabular}%
}
\end{table*}

\subsection{Multi-view Fusion \& Scene-level Regression}
\label{4.4}
Given the sparse view weights, the selected-view evidence is aggregated and regressed as:
\begin{equation}
    \mathbf{h} = \sum_{v=1}^{V} \alpha_v \mathbf{v}_v, \qquad
    \hat{y} = \phi_{\mathrm{reg}}\!\left(\mathrm{LN}\!\left(\mathbf{h} + \phi_{\mathrm{fuse}}(\mathbf{h})\right)\right).
\end{equation}
The prediction is therefore based on the aggregated evidence from selected candidate views, aligning the scoring path with the view-dependent nature of aesthetic assessment.

\subsection{Aesthetics-aware Optimization Objective}

As described in Section~\ref{sec:annotation}, the supervision targets can be the overall scene-level aesthetic score. Given a normalized scene-level target $y$, we optimize the model with a robust regression loss and a pairwise ranking loss:
\begin{equation}
    \mathcal{L}
    =
    \mathcal{L}_{\mathrm{Huber}}(\hat{y}, y)
    +
    \lambda_{\mathrm{rank}}
    \mathcal{L}_{\mathrm{rank}}.
\end{equation}
The Huber term provides a robust objective for fitting the absolute
scene-level score~\citep{huber1992robust}. The ranking term complements it by
encouraging ordinal consistency between scenes. This is important for aesthetic
assessment, where human ratings are naturally ordered rather than purely
categorical. Related image aesthetic assessment methods such as NIMA explicitly
model this ordered nature of human opinion scores by predicting rating
distributions and optimizing an Earth Mover's Distance loss~\citep{IAA11}.
In our scalar scene-level setting, we impose ordinal consistency through a
pairwise ranking objective. Within a mini-batch, valid ranking pairs are formed as
$\mathcal{P}=\{(i,j): |y_i-y_j|>\epsilon\}$. Following the standard pairwise
hinge ranking formulation used in learning-to-rank methods such as Ranking
SVM~\citep{joachims2002optimizing}, we define the ranking loss as:
\begin{equation}
    \mathcal{L}_{\mathrm{rank}}
    =
    \frac{1}{|\mathcal{P}|}
    \sum_{(i,j)\in\mathcal{P}}
    \max
    \left(
        0,
        m
        -
        \mathrm{sign}(y_i-y_j)
        (\hat{y}_i-\hat{y}_j)
    \right).
\end{equation}

\section{Experiments \& Results}
\label{sec:experiment}
\subsection{Experimental Setup}

\noindent\textbf{Datasets and Metrics.}
We conduct experiments on our constructed Aesthetic3D dataset, with 80\% for training and 20\% for testing. For 3D scene input, each scene is represented by 2{,}048 Gaussian primitives obtained via Farthest Point Sampling (FPS) \cite{qi2017pointnet++}. We normalize the geometric coordinates and normalize the scene scale using the 95th-percentile radius. In the multi-view setting, each scene is associated with 32 candidate views selected from real source cameras using uniform azimuth--elevation bins. Since our method directly takes Gaussian primitives as input, it does not rely on rendered 2D images. We report three correlation metrics, \textbf{PLCC}, \textbf{SRCC}, and \textbf{KRCC} ($\uparrow$), together with two calibration metrics, \textbf{MAE} and \textbf{RMSE} ($\downarrow$). Unless stated otherwise, all numbers are reported as $\text{mean}\pm\text{std}$ over run seeds $\{7,13,42\}$. Because split construction is seed-controlled in the current pipeline, these statistics capture both optimization and split variation.

\noindent\textbf{Implementation Details.}
Please refer to the supplementary material for the full implementation details for our method and baselines. Annotation was performed on 2 NVIDIA H100 GPUs, and experiments were performed on a NVIDIA GeForce RTX 5090 GPU. \underline{\emph{Aes3DGSNet}} uses a 4-block point transformer as the scene encoder, projects Gaussian scene tokens onto a $14 \times 14$ view grid through geometric projection for view-level modeling, selects the top-8 most informative views for aggregation, and predicts the scene-level aesthetic score with a 3-layer regressor. We train the model for 16 epochs using AdamW ($\text{lr} = 5 \times 10^{-5}$, $\text{wd} = 10^{-4}$, and batch size $= 4$) with a combination of Huber loss and pairwise ranking loss.  Using the overall score as the supervision label is denoted as (total), while using the average aggregated score over the 8 attributes as the label is denoted as (8-attr). \underline{\emph{(A) Zero-shot 2D Image Aesthetic Baselines}} include ArtiMuse \cite{ASS3}, AADB \cite{ASS2}, NIMA \cite{IAA11}, and TANet \cite{ASS1}. Their pretrained models are used to infer scores for newly rendered views. The scene score is obtained by averaging predictions across views, then applying the same train-split linear calibration used in the evaluation script. \underline{\emph{(B) 3DGS quality assessment baselines}} include MUGSQA-style DBCNN~\cite{baseline4} and GSOQA~\cite{GSP02}. The former follows a rendered-view NR-IQA pipeline by scoring individual views and averaging them, while the latter directly operates on Gaussian primitives for quality prediction. \underline{\emph{(C) Point-based quality assessment baselines}} include MM-PCQA~\cite{baseline1}, LRL-GQA~\cite{baseline2}, and Stochastic NR-PCQA~\cite{baseline3}, for point-based quality assessment.

\subsection{Comparative Experiment}

Table~\ref{tab:main_results} compares Aes3DGSNet with baselines from different paradigms. Aes3DGSNet achieves the best performance with a remarkably lightweight architecture and modest computational cost. The Aes3DGSNet (8-attr) variant achieves the best performance on rank-based correlation metrics, while Aes3DGSNet (total) performs best on MAE and RMSE, indicating that explicit geometry-aware multi-view reasoning is crucial for reliable scene-level aesthetic assessment. In contrast, zero-shot aesthetic priors transfer poorly to 3DGS renderings: ArtiMuse shows weak correlation with the target, even after scene-level aggregation and calibration, revealing a clear domain gap between natural images and rendered views. Deep-AADB, NIMA, and TANet-AVA show similarly weak or negative correlations, further confirming the limited transferability of 2D aesthetic models to 3DGS scenes. Rendered-view no-reference 3DGS quality assessment baseline, MUGSQA-DBCNN, also performs suboptimally, suggesting that image-level quality predictors are insufficient to capture high-level aesthetic attributes in 3D scenes. GSOQA achieves stronger performance due to its geometry-aware design, but still falls short of our method. Point-based quality assessment methods yield moderate results overall. See the supplementary material for qualitative examples and mean/median baselines.

\subsection{Ablation Study}

\definecolor{blockA}{RGB}{255, 240, 240}
\definecolor{blockB}{RGB}{255, 245, 235}
\definecolor{blockC}{RGB}{255, 250, 230}
\definecolor{blockD}{RGB}{240, 250, 240}
\definecolor{blockE}{RGB}{235, 245, 255}
\begin{table}[t]
\centering
\caption{Ablation on the Aes3DGSNet backbone. Best results within each block are in \textbf{bold}.}
\label{tab:ablation}
\setlength{\tabcolsep}{1.5pt}
\resizebox{\linewidth}{!}{%
\begin{tabular}{l l c c c c c}
\toprule
 & Variant & PLCC $\uparrow$ & SRCC $\uparrow$ & KRCC $\uparrow$ & MAE $\downarrow$ & RMSE $\downarrow$ \\
\midrule

\rowcolor{blockA}
\multicolumn{7}{l}{\textit{\textbf{(A) Scene-global token fusion}}}\\
\rowcolor{blockA}
A1 & scene-global token ($\times$) & $.534\mathord{\pm}.016$ & $.507\mathord{\pm}.082$ & $.351\mathord{\pm}.067$ & $.088\mathord{\pm}.003$ & $.115\mathord{\pm}.003$ \\
\rowcolor{blockA}
A2 & Aes3DGSNet & $\mathbf{.567\mathord{\pm}.036}$ & $\mathbf{.554\mathord{\pm}.043}$ & $\mathbf{.389\mathord{\pm}.032}$ & $\mathbf{.084\mathord{\pm}.008}$ & $\mathbf{.110\mathord{\pm}.005}$ \\

\addlinespace[2pt]

\rowcolor{blockB}
\multicolumn{7}{l}{\textit{\textbf{(B) View selector}}}\\
\rowcolor{blockB}
B1 & learnable ctrl. tokens ($\times$) & $.511\mathord{\pm}.019$ & $.493\mathord{\pm}.032$ & $.342\mathord{\pm}.024$ & $.092\mathord{\pm}.008$ & $.118\mathord{\pm}.006$ \\
\rowcolor{blockB}
B2 & uniform (no selector) & $.513\mathord{\pm}.042$ & $.514\mathord{\pm}.059$ & $.369\mathord{\pm}.041$ & $.086\mathord{\pm}.006$ & $.112\mathord{\pm}.009$ \\
\rowcolor{blockB}
B3 & Aes3DGSNet & $\mathbf{.567\mathord{\pm}.036}$ & $\mathbf{.554\mathord{\pm}.043}$ & $\mathbf{.389\mathord{\pm}.032}$ & $\mathbf{.084\mathord{\pm}.008}$ & $\mathbf{.110\mathord{\pm}.005}$ \\

\addlinespace[2pt]

\rowcolor{blockC}
\multicolumn{7}{l}{\textit{\textbf{(C) Top-$K$ sensitivity}}}\\
\rowcolor{blockC}
C1 & $K=8$ (Aes3DGSNet) & $\mathbf{.567\mathord{\pm}.036}$ & $\mathbf{.554\mathord{\pm}.043}$ & $\mathbf{.389\mathord{\pm}.032}$ & $\mathbf{.084\mathord{\pm}.008}$ & $\mathbf{.110\mathord{\pm}.005}$ \\
\rowcolor{blockC}
C2 & $K=16$ & $.545\mathord{\pm}.046$ & $.517\mathord{\pm}.050$ & $.367\mathord{\pm}.039$ & $.085\mathord{\pm}.008$ & $.113\mathord{\pm}.007$ \\
\rowcolor{blockC}
C3 & $K=32$ & $.540\mathord{\pm}.033$ & $.510\mathord{\pm}.045$ & $.366\mathord{\pm}.032$ & $.085\mathord{\pm}.007$ & $.113\mathord{\pm}.007$ \\

\addlinespace[2pt]

\rowcolor{blockD}
\multicolumn{7}{l}{\textit{\textbf{(D) Ranking loss weight}}}\\
\rowcolor{blockD}
D1 & $\mathcal{L}_{\mathrm{rank}}=0.0$ & $.552\mathord{\pm}.008$ & $.518\mathord{\pm}.033$ & $.366\mathord{\pm}.035$ & $.085\mathord{\pm}.006$ & $.111\mathord{\pm}.004$ \\
\rowcolor{blockD}
D2 & $\mathcal{L}_{\mathrm{rank}}=0.1$ (Aes3DGSNet) & $\mathbf{.567\mathord{\pm}.036}$ & $\mathbf{.554\mathord{\pm}.043}$ & $\mathbf{.389\mathord{\pm}.032}$ & $\mathbf{.084\mathord{\pm}.008}$ & $\mathbf{.110\mathord{\pm}.005}$ \\
\rowcolor{blockD}
D3 & $\mathcal{L}_{\mathrm{rank}}=0.5$ & $.548\mathord{\pm}.026$ & $.533\mathord{\pm}.049$ & $.374\mathord{\pm}.043$ & $.087\mathord{\pm}.009$ & $.112\mathord{\pm}.007$ \\

\addlinespace[2pt]

\rowcolor{blockE}
\multicolumn{7}{l}{\textit{\textbf{(E) Geometric projection}}}\\
\rowcolor{blockE}
E1 & geometric projection ($\times$) & $.398\mathord{\pm}.090$ & $.385\mathord{\pm}.099$ & $.274\mathord{\pm}.074$ & $.094\mathord{\pm}.011$ & $.120\mathord{\pm}.012$ \\
\rowcolor{blockE}
E2 & Aes3DGSNet & $\mathbf{.567\mathord{\pm}.036}$ & $\mathbf{.554\mathord{\pm}.043}$ & $\mathbf{.389\mathord{\pm}.032}$ & $\mathbf{.084\mathord{\pm}.008}$ & $\mathbf{.110\mathord{\pm}.005}$ \\

\bottomrule
\end{tabular}%
}
\end{table}

Table~\ref{tab:ablation} isolates the contribution of each design choice on the Aes3DGSNet backbone, and more ablations can be found in the supplementary material. The differences mainly reflect where information enters the model and how the view evidence is selected and pooled. \underline{\emph{(A) Scene-global token fusion:}} Removing the scene-global token (A1) leads to consistent degradation across all metrics relative to the full model (A2), showing that global scene context is beneficial for effective view selection. \underline{\emph{(B) View selector:}} Both simplified selector variants underperform the full model. In particular, removing learnable control tokens (B1) yields the most noticeable drop, while uniform aggregation without a selector (B2) also weakens performance. This verifies the importance of learned view selection for aggregating informative view evidence. \underline{\emph{(C) Top-$K$ sensitivity:}} Increasing $K$ from $8$ to $16$ or $32$ consistently weakens rank metrics, suggesting that sparse view selection acts as a regularizer by focusing on the most informative views. \underline{\emph{(D) Ranking loss weight:}} Disabling the ranking term (D1) leaves PLCC/RMSE largely unchanged but reduces SRCC/KRCC, indicating that regression alone is insufficient for rank fidelity. Increasing $\mathcal{L}_{\mathrm{rank}}$ to $0.5$ (D3) does not yield further gains and harms calibration. $\mathcal{L}_{\mathrm{rank}}{=}0.1$ provides the best trade-off. \underline{\emph{(E) Geometric projection:}} This setup using only the scene encoder without geometric projection (E1) performs substantially worse across all metrics, indicating that view-conditioned projected evidence provides layout, visibility, and spatial-configuration cues that cannot be captured by a single scene-global representation alone.

\definecolor{blockF}{RGB}{245, 240, 255}
\definecolor{blockG}{RGB}{235, 245, 255}
\begin{table}[t]
\centering
\caption{Gaussian count and candidate-view overlap ablations. Values are reported point estimates for each setting. Table~\ref{tab:main_results} reports mean$\pm$std over three seeds. Best values per panel are in \textbf{bold}.}
\label{tab:budget_overlap_ablation}
\setlength{\tabcolsep}{2pt}
\resizebox{\linewidth}{!}{%
\begin{tabular}{l c c c c c c}
\toprule
$N$ & PLCC $\uparrow$ & SRCC $\uparrow$ & KRCC $\uparrow$ & MAE $\downarrow$ & RMSE $\downarrow$ & GFLOPs $\downarrow$ \\
\midrule
\rowcolor{blockF}
\multicolumn{7}{l}{\textit{\textbf{(F) Gaussian count}}} \\
\rowcolor{blockF} 1,024 & $.470$ & $.480$ & $.338$ & $.098$ & $.115$ & $\mathbf{14.8}$ \\
\rowcolor{blockF} 2,048 (default) & $.567$ & $.554$ & $.389$ & $.084$ & $.110$ & $29.4$ \\
\rowcolor{blockF} 4,096 & $\mathbf{.587}$ & $\mathbf{.567}$ & $\mathbf{.407}$ & $\mathbf{.083}$ & $\mathbf{.104}$ & $81.2$ \\
\bottomrule
\end{tabular}%
}
\par\vspace{1pt}
\setlength{\tabcolsep}{3pt}
\resizebox{\linewidth}{!}{%
\begin{tabular}{l c c c c}
\toprule
Probe setting & Overlap & PLCC $\uparrow$ & SRCC $\uparrow$ & Prediction-rank consistency $\uparrow$ \\
\midrule
\rowcolor{blockG}
\multicolumn{5}{l}{\textit{\textbf{(G) Candidate-view overlap}}} \\
\rowcolor{blockG} Default & $10/32$ & $\mathbf{.557}$ & $.584$ & $\mathbf{1.000}$ \\
\rowcolor{blockG} Full overlap & $32/32$ & $.538$ & $\mathbf{.591}$ & $.989$ \\
\rowcolor{blockG} Disjoint & $0/32$ & $.540$ & $.589$ & $.991$ \\
\rowcolor{blockG} Far-disjoint & $0/32$ & $.508$ & $.551$ & $.944$ \\
\bottomrule
\end{tabular}%
}
\end{table}

\noindent\underline{\emph{(F) Gaussian count:}} Table~\ref{tab:budget_overlap_ablation} shows that increasing $N$ from 1,024 to 2,048 substantially improves correlation, while 4,096 gives smaller gains at 2.8$\times$ the computation of 2,048. We use 2,048 as a practical accuracy--cost trade-off. \underline{\emph{(G) Candidate-view overlap:}} Full-overlap and disjoint probes give similar PLCC/SRCC to the default setting. Far-disjoint probes reduce both metrics, but their prediction-rank consistency remains $.944$.

\section{Conclusion}
We present Aes3D, the first systematic framework for aesthetic assessment of 3D neural rendering scenes. Our framework includes Aesthetic3D, the first 3D scene aesthetic dataset, and Aes3DGSNet, a lightweight, rendering-independent model that predicts 3DGS scene-level aesthetic scores directly from Gaussian primitives. Projection-based view tokenization and learned view selection capture view-dependent aesthetic cues without rendering RGB images. Experimental results demonstrate that Aes3DGSNet achieves competitive aesthetic assessment performance with high efficiency, establishing a strong benchmark for this new task. We hope this work will encourage further research on 3D scene aesthetics and support aesthetics-aware content creation and editing.

{\small
\bibliographystyle{ieeenat_fullname}
\bibliography{references}
}

\clearpage
\appendix
\setcounter{figure}{0}
\renewcommand{\thefigure}{S\arabic{figure}}
\setcounter{table}{0}
\renewcommand{\thetable}{S\arabic{table}}
\setcounter{equation}{0}
\renewcommand{\theequation}{S\arabic{equation}}
\definecolor{blockA}{RGB}{255,240,240}
\definecolor{blockB}{RGB}{255,245,235}
\definecolor{blockC}{RGB}{255,250,230}
\definecolor{blockD}{RGB}{240,250,240}
\definecolor{blockE}{RGB}{235,245,255}
\twocolumn[{\centering\Large\bfseries Supplementary Material\par\vspace{1em}}]
\section{Broader Impact}
\label{appendix:broader_impact}
Aes3D may support aesthetics-aware 3D content creation, quality control, rendering optimization, and digital media production by providing automatic scene-level aesthetic feedback for 3DGS content. This can reduce the cost of evaluating large-scale 3D assets and help creators identify visually weak scenes or viewpoints.

Potential negative impacts include over-reliance on automated aesthetic scores, homogenization of visual styles, and biases inherited from the MLLM-based annotator and source datasets. Since aesthetic preference is subjective and culturally dependent, Aes3D should assist evaluation rather than define artistic value.

\section{Asset Licenses}
\label{appendix:Asset Licenses}
We use DL3DV-10K and Bilarf under their respective research licenses/terms of use, and cite the original papers. ArtiMuse is used as an external annotator following its released usage terms. Our released annotations and code will follow licenses compatible with the original assets.

\section{Reasons for Selecting the 3D Scene Datasets}
\label{IAA-Dataset}

Selecting appropriate 3D scene datasets is critical for constructing a reliable benchmark for 3D aesthetic assessment.
Unlike conventional 3D reconstruction or perceptual quality evaluation, our task requires datasets that not only provide accurate geometry and appearance, but also exhibit diverse aesthetic attributes under realistic viewing conditions.
Therefore, we select DL3DV-10K and Bilarf as complementary data sources, considering their scene diversity, acquisition characteristics, and coverage of challenging visual conditions.
These choices also fit our annotation pipeline, which aggregates multi-view 2D aesthetic scores into scene-level labels.

\noindent\textbf{DL3DV-10K (CVPR 2024)} \cite{DT2}:
DL3DV-10K is a large-scale real-world dataset containing approximately 10,000 scenes, covering both indoor and outdoor environments.
A key advantage of this dataset lies in its data acquisition paradigm: scenes are captured using consumer-grade devices (e.g., mobile phones), rather than controlled capture setups or synthetic pipelines.
Thus, the dataset reflects user-generated scenes with varied composition, viewpoints, object arrangements, and scene semantics.

From the perspective of 3D aesthetic assessment, DL3DV-10K provides several important properties. First, its large-scale and diverse scene coverage enables the dataset to span a wide range of aesthetic qualities, avoiding bias toward specific scene categories or styles.
This is essential for learning a robust mapping between 3D structure and aesthetic perception. Second, the dataset naturally contains human-centric and everyday scenes (e.g., indoor layouts, urban environments, and daily activities), which are closely aligned with real-world aesthetic evaluation scenarios. Aesthetic perception is strongly influenced by semantic content and spatial composition rather than purely geometric fidelity. Third, DL3DV-10K provides multi-view image observations suitable for constructing viewpoint-dependent aesthetic annotations.
Since 3D scene aesthetics emerges from aggregating perceptions across views, such dense multi-view coverage is crucial for capturing intra-scene aesthetic variation and supporting our multi-view aggregation pipeline.

Overall, DL3DV-10K is our primary source of diverse real-world 3D scenes for aesthetic modeling.

\noindent\textbf{Bilarf (TOG 2024)} \cite{DT3}:
In contrast to DL3DV-10K, Bilarf focuses on more challenging real-world imaging conditions, particularly scenarios involving low-light, night-time environments, and complex illumination variations.
It adds lighting, contrast, and color variation relevant to aesthetic judgments.

From an aesthetic assessment perspective, Bilarf provides complementary benefits. First, lighting is one of the most influential factors in human aesthetic judgment, affecting mood, visibility, contrast, and color harmony. The inclusion of low-light and high-dynamic-range conditions allows the benchmark to evaluate whether models can capture aesthetic variations beyond well-lit canonical scenes. Second, challenging illumination conditions often amplify rendering artifacts in 3DGS (e.g., noise, color inconsistency, or over-smoothing), which may not be reflected in standard perceptual metrics but can significantly impact aesthetic quality.
Therefore, Bilarf enables us to test the robustness of aesthetic assessment under non-ideal rendering conditions. Third, Bilarf adds a distribution shift that helps prevent overfitting to common scene statistics and improves generalization.

\noindent\textbf{Complementarity and Benchmark Design.}
The combination of DL3DV-10K and Bilarf is designed to cover both:
(i) large-scale, natural, and semantically rich scenes, and
(ii) visually challenging conditions with strong aesthetic variability.

This complementary design aligns with our goal of constructing a general-purpose benchmark for 3D scene aesthetic assessment.
Existing evaluation datasets for 3DGS mainly focus on perceptual fidelity (e.g., distortion, sharpness), while largely ignoring higher-level aesthetic attributes such as composition, lighting, and emotional impact.

By integrating these two datasets and annotating them using a unified aesthetic scoring pipeline, we obtain a dataset that:
\begin{itemize}
    \item spans diverse scene categories and aesthetic styles,
    \item captures strong viewpoint-dependent aesthetic variation,
    \item includes both ideal and non-ideal imaging conditions,
    \item and better reflects real-world use cases in immersive media and content creation.
\end{itemize}

Therefore, the selected datasets provide a balanced and representative foundation for evaluating 3D scene aesthetics in the context of 3D Gaussian Splatting.

\noindent\textbf{Why not Other Datasets like NeRF Synthetic / Tanks \& Temples.}
We intentionally do not adopt commonly used datasets such as NeRF Synthetic \cite{mildenhall2020nerf} or Tanks \& Temples \cite{knapitsch2017tanks}, as they are not well aligned with the objective of 3D aesthetic assessment. NeRF Synthetic consists of computer-generated scenes with highly controlled geometry, lighting, and texture distributions, which differ significantly from real-world visual statistics. While suitable for evaluating reconstruction fidelity, such synthetic data lacks the semantic richness and natural variability that are essential for modeling human aesthetic perception, which is strongly influenced by composition, scene semantics, and contextual cues. In addition, both NeRF Synthetic and Tanks \& Temples provide limited support for viewpoint-dependent aesthetic modeling. NeRF Synthetic typically exhibits simple scene layouts and regular camera trajectories, resulting in weak variation of aesthetic perception across views. Tanks \& Temples, although based on real captures, is relatively small-scale and focuses on geometry-centric benchmarks with limited scene diversity. More importantly, these datasets are designed for evaluating reconstruction accuracy rather than high-level attributes such as visual appeal, lighting aesthetics, and emotional expression. Therefore, we instead adopt DL3DV-10K and Bilarf, which better reflect real-world content distributions and provide richer conditions for 3D aesthetic assessment.

\noindent\textbf{Dataset Limitations.}
Despite the careful design of Aesthetic3D, several limitations remain. First, the dataset size is still relatively limited at the scene level, which may constrain generalization to long-tail scenarios. Second, aesthetic annotations are obtained via an MLLM-based scoring pipeline rather than large-scale human studies. Although this approach is scalable and consistent, it may introduce biases inherited from the underlying model and may not fully capture subjective human preferences. Third, the dataset primarily covers everyday real-world environments, with limited representation of highly stylized or artistic scenes such as cinematic or game content. In addition, while Bilarf introduces challenging illumination conditions, extreme visual scenarios are still underrepresented. Finally, the current formulation focuses on static multi-view aggregation and does not explicitly model temporal dynamics or interactive viewing, which may be important for video-based or immersive aesthetic assessment. We leave these directions for future work.

\section{Comparison between IAA Annotators}
\label{appendix:Comparison between IAA Annotators}

\begin{figure*}[t]
    \centering
    \includegraphics[width=\textwidth]{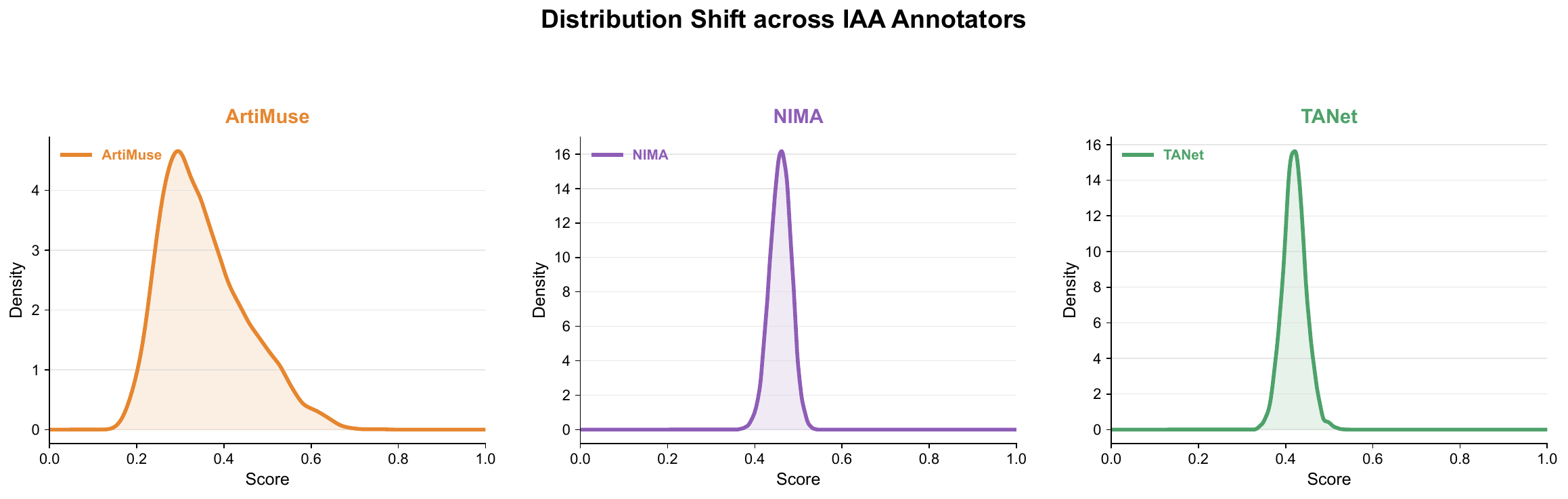}
    \caption{Score distributions of different IAA annotators on Aesthetic3D Dataset. ArtiMuse shows a broader score range, capturing richer aesthetic variation across scenes, while NIMA and TANet produce highly concentrated distributions with limited discrimination.}
    \label{fig:iaa_distribution}
\end{figure*}

The distribution analysis further supports our choice of ArtiMuse as the supervision signal for Aesthetic3D.
As shown in Figure~\ref{fig:iaa_distribution}, ArtiMuse exhibits a significantly broader and more expressive score distribution compared with NIMA \cite{IAA11} and TANet \cite{ASS1}.
This indicates that ArtiMuse preserves meaningful aesthetic variation across 3D scenes, rather than collapsing most samples into a narrow interval.

\noindent\textbf{Limitations of NIMA and TANet.}
NIMA~\cite{IAA11} formulates image aesthetic assessment as a distribution prediction problem, where a convolutional neural network predicts a histogram over aesthetic scores and derives the mean score as the final output.
Although effective for natural images, NIMA is primarily trained on datasets such as AVA, where aesthetics are largely driven by photographic composition and appearance statistics.
As a result, it tends to produce conservative predictions and exhibits limited dynamic range when applied to out-of-domain data.

TANet~\cite{ASS1} further improves aesthetic assessment by incorporating attribute-aware attention mechanisms, aiming to capture multiple aesthetic factors in a unified framework.
However, similar to NIMA, TANet is still fundamentally designed for 2D photographic aesthetics and relies on image-level cues.
When applied to rendered views of 3DGS scenes, both NIMA and TANet show strong distribution collapse, with most scores concentrated in a narrow band (as observed in Figure~\ref{fig:iaa_distribution}).
Such compressed distributions provide weak supervision signals, making it difficult for regression models to distinguish between subtle differences in scene-level aesthetic quality.

\noindent\textbf{Advantages of ArtiMuse.}
In contrast, ArtiMuse is built upon a multimodal large language model with explicit fine-grained aesthetic reasoning capabilities.
Instead of predicting a single holistic score directly, it decomposes aesthetic quality into multiple interpretable attributes, including composition, visual structure, technical execution, originality, theme communication, emotional response, gestalt, and overall quality.
This attribute-level evaluation aligns well with the multi-factor nature of 3D scene aesthetics, where spatial arrangement, structural completeness, and semantic coherence play a critical role.

More importantly, the resulting score distribution of ArtiMuse spans a much wider range.
This larger dynamic range implies higher sensitivity to aesthetic variations and leads to stronger ranking signals across scenes.
From a learning perspective, such diversity is crucial: it improves regression stability, enhances correlation learning (PLCC/SRCC), and avoids the gradient saturation problem caused by overly concentrated targets.
In other words, ArtiMuse provides a more informative supervision signal that better reflects relative aesthetic ordering among 3D scenes.

\noindent\textbf{Right-skewed Distribution.}
We also observe that the ArtiMuse distribution is slightly right-skewed, with more samples receiving mid-to-low scores.
This should not be interpreted as a weakness of the annotator, but rather as a reflection of the intrinsic characteristics of the datasets used.
Both DL3DV-10K and BiLARF are designed for general-purpose 3D reconstruction rather than aesthetic optimization.
They contain a large number of casually captured, real-world scenes with unconstrained viewpoints, imperfect lighting conditions, and non-ideal compositions.
Therefore, it is expected that many scenes do not exhibit high aesthetic quality under a rigorous evaluation criterion.

From a dataset construction perspective, this skewness is in fact desirable.
It indicates that ArtiMuse is capable of making strict and discriminative judgments, avoiding artificial score inflation.
At the same time, the presence of a long tail toward higher scores ensures that high-quality scenes are still distinguishable, enabling effective learning of both absolute quality and relative ranking.

\noindent\textbf{Summary.}
Overall, compared with NIMA and TANet, ArtiMuse provides (1) a significantly broader and more discriminative score distribution, (2) semantically grounded attribute-level supervision aligned with 3D scene aesthetics, and (3) realistic scoring behavior consistent with the nature of real-world 3D datasets.
These properties make it a more suitable and reliable IAA annotator for Aesthetic3D.

\section{3D Gaussian Reconstruction from Multi-view Images}
\label{sec:appendix_3dgs_reconstruction}
\subsection{Reconstruction Method - DepthSplat}
To construct the 3D Gaussian scenes used in our benchmark, we follow \textbf{DepthSplat}~\cite{xu2024depthsplat}, a feed-forward sparse-view 3D reconstruction framework that connects multi-view depth estimation and 3D Gaussian Splatting. Given a set of posed multi-view images, DepthSplat first predicts geometrically consistent depth maps and then back-projects them into 3D space to obtain Gaussian centers. A lightweight decoder is further used to predict the remaining Gaussian attributes, such as appearance, opacity, scale, and orientation, which together form the final 3D Gaussian representation.

A key component of DepthSplat is that it augments multi-view matching with features from a pre-trained monocular depth model. This improves depth estimation in challenging regions, such as texture-less areas, occlusions, and reflective surfaces, while still maintaining cross-view geometric consistency. The resulting depth maps provide a reliable geometric scaffold for Gaussian reconstruction.

Formally, for a pixel $\mathbf{u}$ with predicted depth $d_i(\mathbf{u})$ in view $i$, the corresponding 3D point is obtained by back-projection:
$$
\mathbf{x}_{i}(\mathbf{u}) = d_i(\mathbf{u}) K_i^{-1} \tilde{\mathbf{u}},
$$
and transformed to the world coordinate system by:
$$
\mathbf{X}_{i}(\mathbf{u}) = T_i \mathbf{x}_{i}(\mathbf{u}),
$$
where $K_i$ and $T_i$ denote the camera intrinsics and extrinsics, respectively. These 3D points are used as Gaussian centers, while the remaining Gaussian parameters are predicted by the decoder and rendered through differentiable Gaussian splatting.

Aes3DGSNet then assesses these 3DGS scenes.

\subsection{3DGS Implementation Details}
For 3D reconstruction and rendering, we use the official \textbf{DepthSplat} implementation with its sparse-view evaluation setting. Each scene is represented by a set of posed multi-view RGB images together with the corresponding camera intrinsics and extrinsics. Before inference, the scene data are converted into the input format required by DepthSplat, including image content, timestamps, normalized camera parameters, and world-to-camera poses.

We perform inference using a publicly released pretrained DepthSplat checkpoint trained for sparse-view 3D Gaussian reconstruction. All input images are resized to a resolution of $270 \times 480$. For each scene, we select 32 context views as reconstruction input. These views are chosen by farthest-point sampling over camera centers, initialized from the camera closest to the scene center, in order to maximize viewpoint coverage and reduce redundancy. Based on the selected context views, we further render 8 candidate target views per scene for inspection and visualization.

The main inference configuration follows a low-resolution feed-forward setting, with an upsampling factor of 8, a lowest feature resolution of 8, a maximum Gaussian scale of 0.1, and a render chunk size of 8. We process every test frame without interval skipping. During inference, we save only the rendered RGB outputs, while disabling auxiliary outputs such as Gaussian export, score computation, ground-truth image saving, and input image saving.

For the final visualization set, the best rendered result is selected from the candidate views using a simple image-quality heuristic based on fill ratio, luminance, contrast, sharpness, and border black ratio. The selected image is resized to $1600 \times 900$ for visualization and analysis.

\section{Full Analysis on Annotation Statistics}
\label{appendix:Full Analysis on Annotation Statistics}
\subsection{Scene Categories}
The source datasets do not share scene category labels, so we grouped all 278 scenes by their dominant content, using the DL3DV-10K taxonomy as a guide. One author assigned the categories, a second reviewed every assignment, and ambiguous cases were resolved by discussion. Table~\ref{tab:scene_categories} reports the five broad groups. These labels do not describe all within-category differences in visual style.

\begin{table}[t]
\centering
\small
\caption{Broad scene categories in Aesthetic3D.}
\label{tab:scene_categories}
\setlength{\tabcolsep}{4pt}
\begin{tabularx}{\linewidth}{Xrr}
\toprule
Scene category & Scenes & Share \\
\midrule
\rowcolor{blockA}
Retail Environments & 69 & 24.8\% \\
\rowcolor{blockB}
Urban Streetscapes & 63 & 22.7\% \\
\rowcolor{blockC}
Landmark Sites & 65 & 23.4\% \\
\rowcolor{blockD}
Workplace and Service Environments & 45 & 16.2\% \\
\rowcolor{blockE}
Natural Landscapes & 36 & 12.9\% \\
\midrule
Total & 278 & 100.0\% \\
\bottomrule
\end{tabularx}
\end{table}

\subsection{Additional Analysis on Total Aesthetic Scores}
\label{app:total_score_stats}

\begin{figure*}[t]
    \centering
    \includegraphics[width=0.8\textwidth]{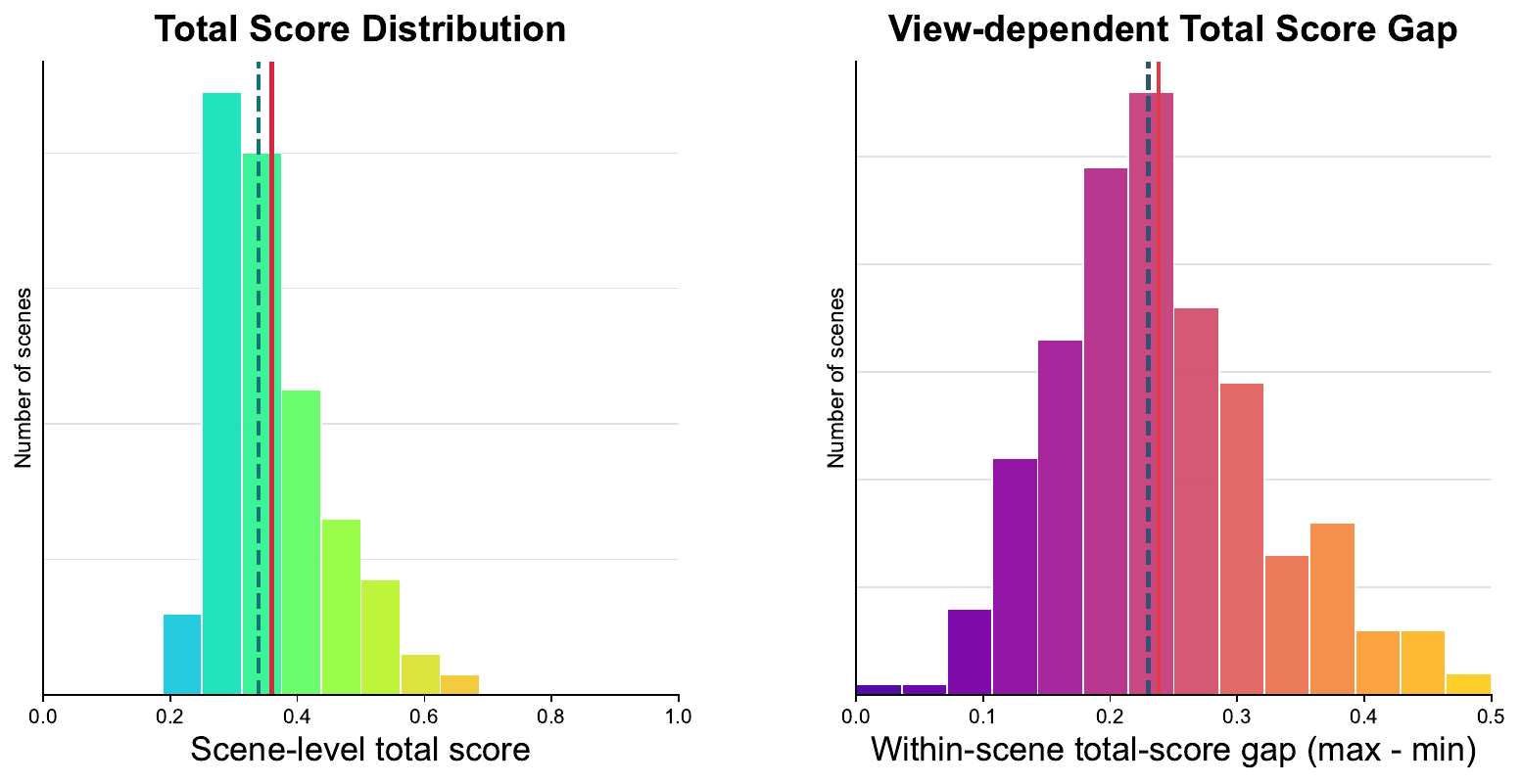}
    \caption{
    Distribution of scene-level total scores (left) and within-scene total-score gaps (right).
    The total-score gap is defined as the difference between the maximum and minimum view-level
    scores within each scene.
    }
    \label{fig:total_score_statistics}
\end{figure*}

In addition to the main analysis based on the 8-attribute mean scores presented in the main paper,
we provide a complementary analysis of the total aesthetic scores derived from ArtiMuse.
Figure~\ref{fig:total_score_statistics} summarizes the distribution of scene-level total scores
and the within-scene view-dependent score variation.

Aesthetic3D contains 278 reconstructed scenes and 92,649 reference views, averaging 333.3 views
per scene. The scene-level total aesthetic score exhibits a non-degenerate distribution, spanning from
0.196 to 0.662, with a mean of 0.359, a median of 0.339, and a standard deviation of 0.092. This shows that the dataset covers a meaningful range of aesthetic quality instead of a narrow interval.

More importantly, we observe strong view dependency in 3D scene aesthetics. For each scene, we
measure the within-scene score gap as the difference between the maximum and minimum view-level
total scores. The resulting gap has a mean of 0.238, a median of 0.230, a 90th percentile of 0.361,
and a maximum of 0.499. Moreover, 86.3\% of scenes have a score gap larger than 0.15, and
67.3\% exceed 0.20.

These statistics confirm that aesthetic perception varies substantially across viewpoints, further
motivating a multi-view assessment paradigm rather than single-view prediction. They also show
that the total-score formulation leads to conclusions consistent with the 8-attribute mean analysis
reported in the main paper.

\subsection{Additional Analysis}
\begin{figure*}[t]
\centering
\includegraphics[width=0.8\textwidth]{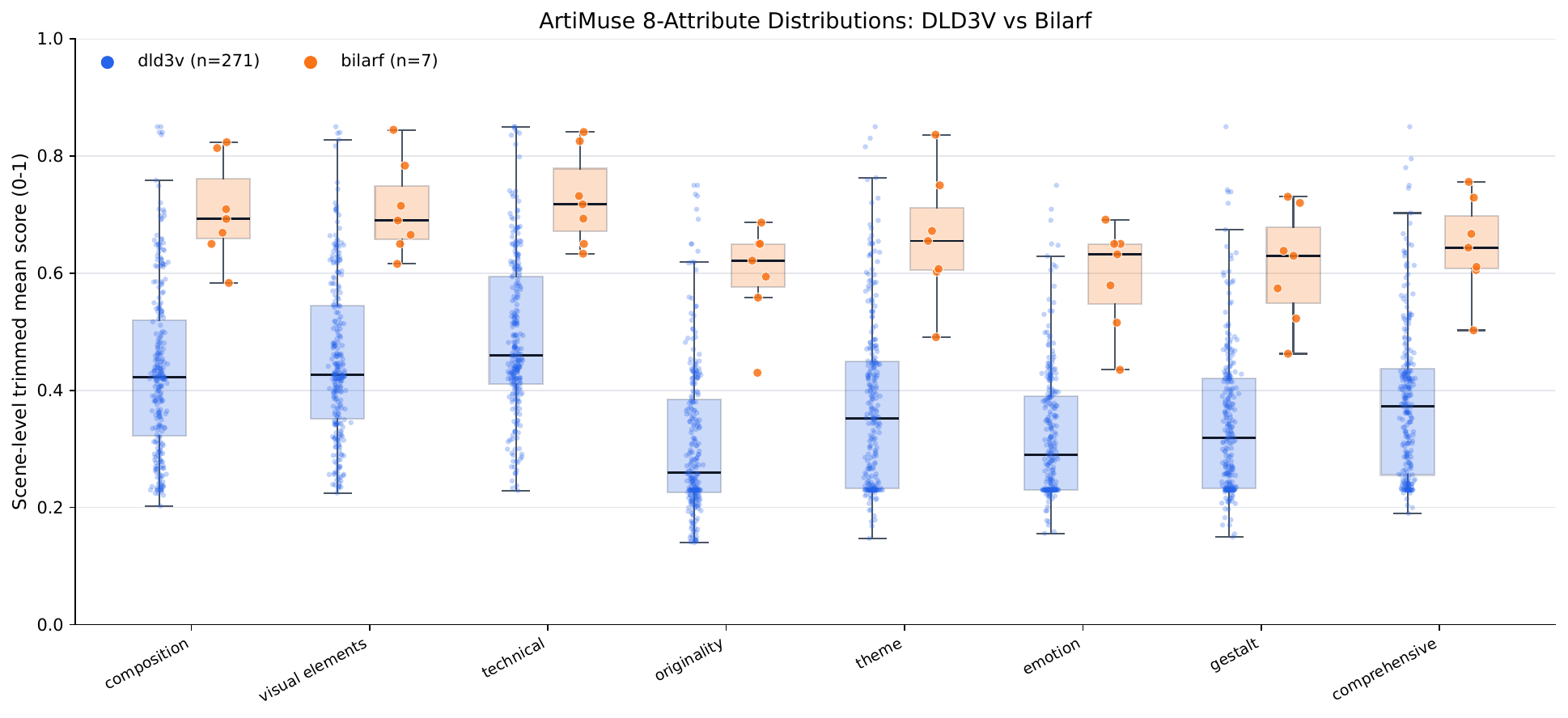}
\caption{
\textbf{Distribution of ArtiMuse attribute scores across datasets.}
Box plots of the eight attribute-level scores for DL3DV-10K and Bilarf.
Bilarf exhibits consistently higher score ranges across all attributes, indicating a clear distribution shift toward higher aesthetic quality.
Due to the small sample size of Bilarf, the statistics mainly reveal the existence of dataset bias rather than a stable estimate of its distribution.
}
\label{fig:attr_dist}
\end{figure*}

\begin{figure*}[t]
\centering
\begin{minipage}{0.4\textwidth}
    \centering
    \includegraphics[width=\linewidth]{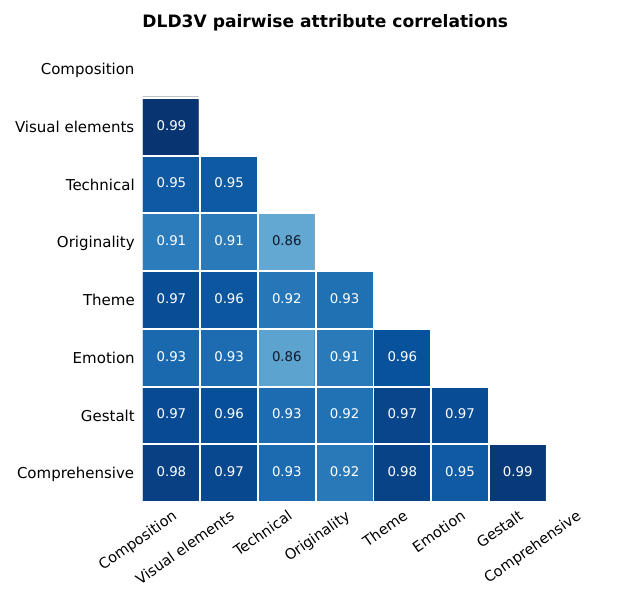}
    \subcaption{DL3DV-10K}
\end{minipage}
\hfill
\begin{minipage}{0.4\textwidth}
    \centering
    \includegraphics[width=\linewidth]{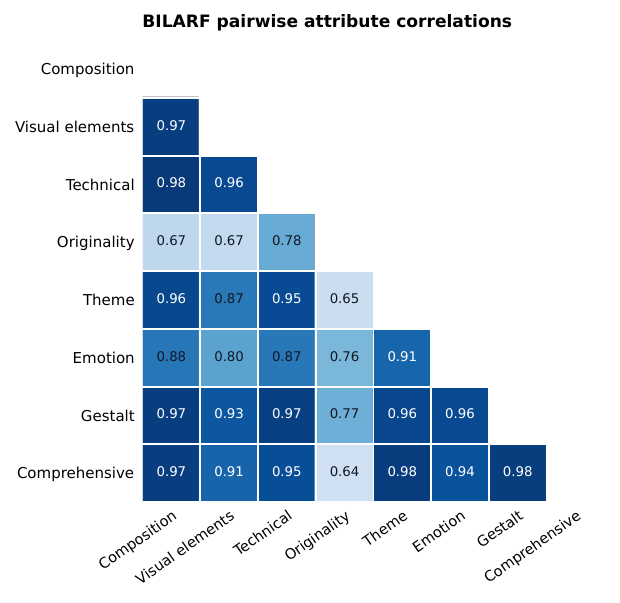}
    \subcaption{Bilarf}
\end{minipage}

\caption{
\textbf{Pairwise Pearson correlations among the eight aesthetic attributes.}
Left: DL3DV-10K shows consistently high correlations (0.86–0.99), indicating strong collinearity among attributes.
Right: Bilarf exhibits similar trends, but originality shows weaker correlations with other attributes, suggesting partial independence under certain conditions.
}
\label{fig:attr_corr}
\end{figure*}

\begin{figure}[t]
\centering
\includegraphics[width=\linewidth]{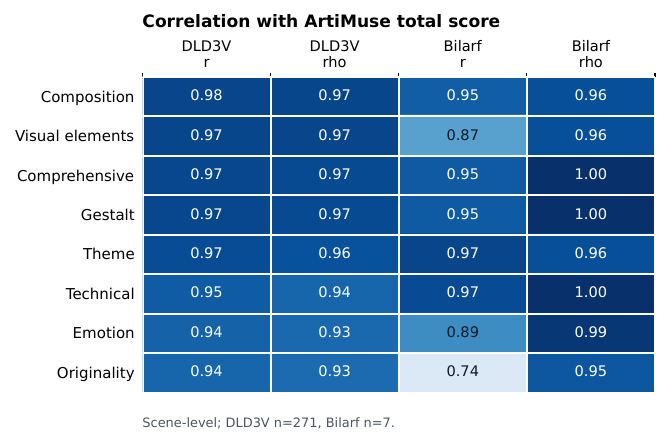}
\caption{
\textbf{Correlation between attribute-level scores and the holistic aesthetic score.}
On DL3DV-10K, all attributes show very strong correlations with the total score (Pearson $\sim0.94$–$0.98$).
On Bilarf, ranking consistency remains high (Spearman $\approx1.0$), while linear correlations vary more, especially for originality.
This indicates strong ordinal consistency but dataset-dependent linear relationships.
}
\label{fig:attr_total_corr}
\end{figure}

In this section, we provide a comprehensive analysis of the statistical properties of the Aesthetic3D annotations. We focus on three aspects: (1) cross-dataset distribution characteristics, (2) inter-attribute relationships, and (3) the consistency between attribute-level scores and the overall aesthetic score. These analyses aim to validate the reliability of the constructed supervision and reveal key properties of the annotation pipeline.

\noindent\textbf{Cross-dataset Distribution Shift.}
We first examine the distribution of the eight attribute-level scores across the two source datasets, DL3DV-10K (denoted as DL3DV-10K) and Bilarf. As illustrated in Figure~\ref{fig:attr_dist}, a distribution shift is observed between the two datasets.

On DL3DV-10K (n=271), most attribute scores are concentrated in the range of approximately $[0.3, 0.5]$, indicating moderate aesthetic quality across diverse real-world scenes. In contrast, Bilarf (n=7) exhibits consistently higher scores across all attributes, with medians typically around $[0.6, 0.73]$.

This result suggests that Bilarf does not follow the same distribution as DL3DV-10K, but instead represents a subset with systematically higher aesthetic quality. We note that the number of Bilarf scenes is relatively small, and therefore the observed statistics should be interpreted as evidence of a distribution shift rather than a stable estimate of its underlying distribution. This discrepancy highlights the presence of dataset bias, which should be considered when designing and evaluating 3D aesthetic assessment models.

\noindent\textbf{Inter-attribute Correlations.}
We next analyze the pairwise Pearson correlations among the eight aesthetic attributes. On DL3DV-10K, the correlations are consistently high, ranging from approximately $0.86$ to $0.99$ (see Figure~\ref{fig:attr_corr}). For example, composition and visual elements reach a correlation of $0.99$, while gestalt and comprehensive evaluation also approach $0.99$. Even the lowest correlations (e.g., involving technical or originality) remain above $0.86$.

These results indicate strong collinearity among the attributes. The eight dimensions largely reflect a shared aesthetic signal rather than independent factors.

On Bilarf, most attribute pairs still exhibit high correlations (often above $0.9$), particularly among composition, technical, theme, gestalt, and comprehensive evaluation. However, originality shows noticeably weaker correlations with other attributes, such as $0.64$ with comprehensive and $0.65$ with theme. This suggests that originality may behave as a relatively more independent dimension under certain conditions. Again, this observation is influenced by the small sample size (n=7) and should be interpreted with caution.

\noindent\textbf{Correlation with Overall Aesthetic Score.}
To further evaluate the validity of attribute-level supervision, we compute the correlation between each attribute and the ArtiMuse holistic aesthetic score, referring to Figure~\ref{fig:attr_total_corr}. On DL3DV-10K, all attributes show very strong correlations with the total score, with Pearson coefficients in the range of approximately $0.94$ to $0.98$ and similarly high Spearman correlations. Composition achieves the highest correlation ($r \approx 0.98$), while even the lowest (originality) remains above $0.94$.

On Bilarf, Spearman correlations are consistently close to $1.0$, indicating strong agreement in ranking. However, Pearson correlations show larger variation: while theme and technical remain high ($\sim0.97$), originality drops to around $0.74$, and visual elements and emotion are also relatively lower.

These results suggest that attribute-level scores preserve strong ordinal consistency with the overall aesthetic judgment, while their linear relationships may become less stable under distribution shift and limited sample conditions.
\begin{figure*}[t]
    \centering
    \includegraphics[width=0.8\textwidth]{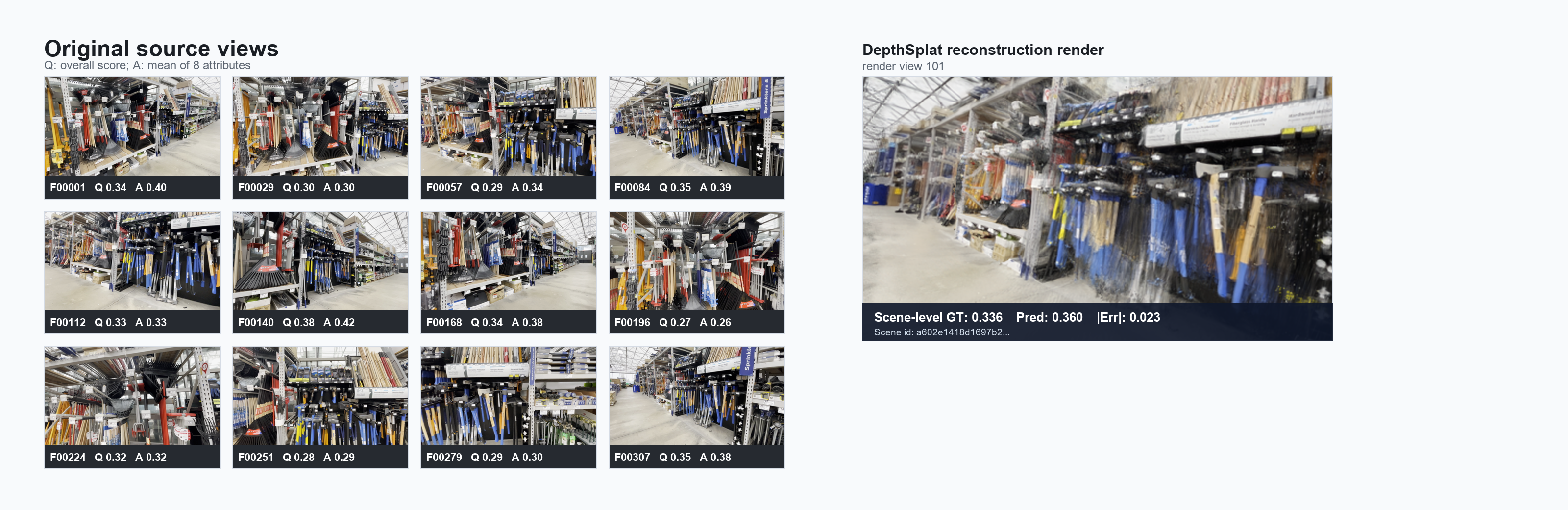}
    \vspace{-2.3em}

    \includegraphics[width=0.8\textwidth]{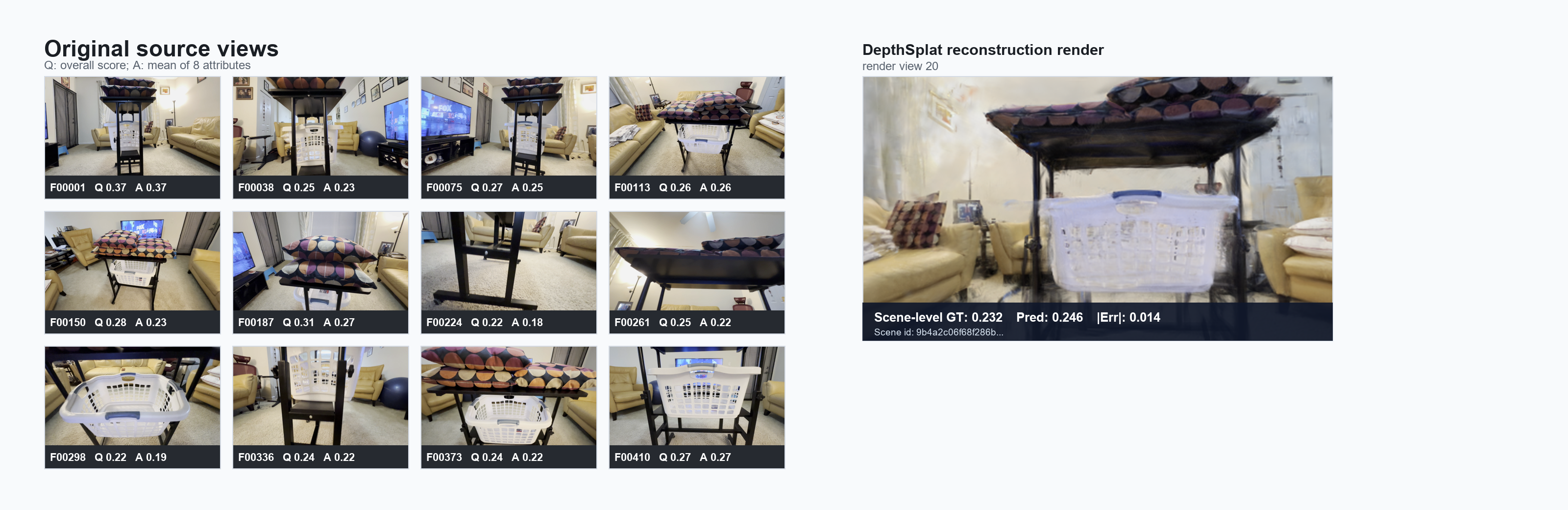}
    \vspace{-2.3em}

    \includegraphics[width=0.8\textwidth]{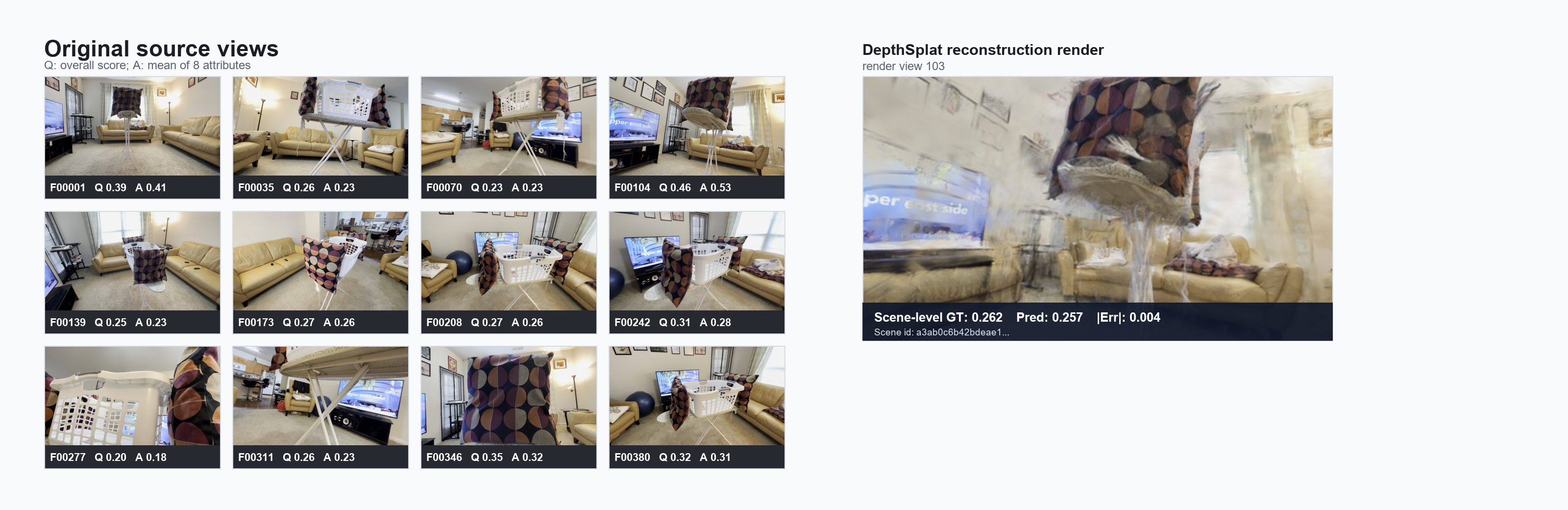}

    \caption{Visualizations of Data Annotation Examples (1).}
    \label{fig:data_examples_vertical}
\end{figure*}

\begin{figure*}[t]
    \centering
    \includegraphics[width=0.8\textwidth]{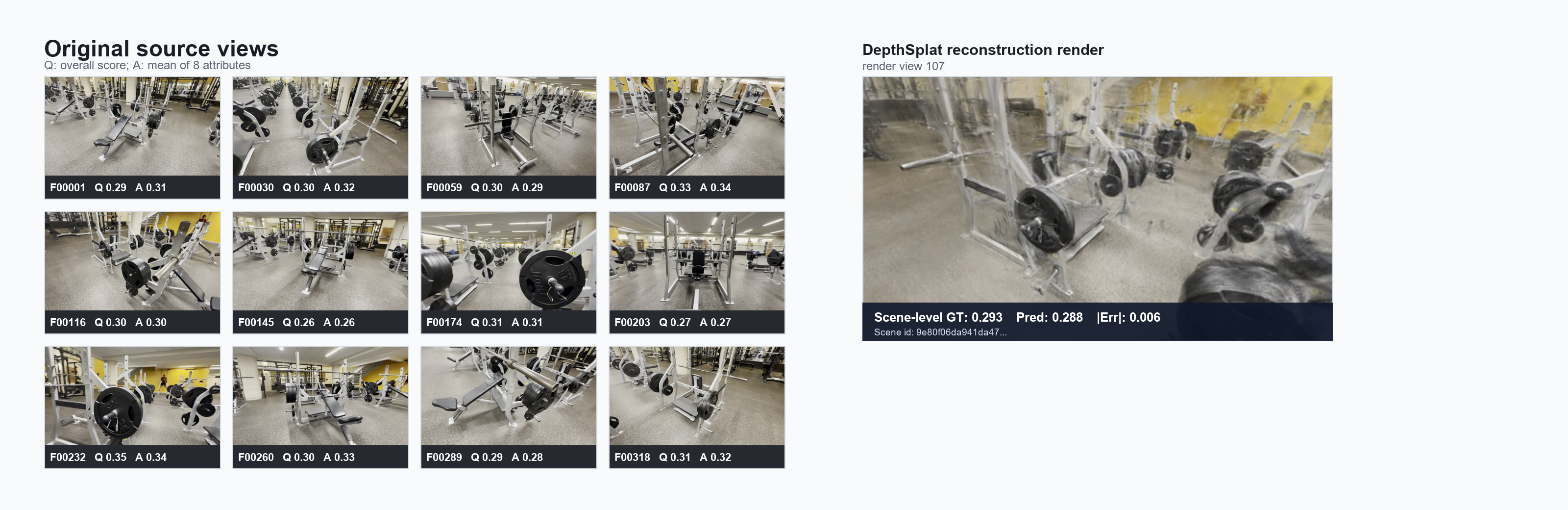}
    \vspace{-2.3em}

    \includegraphics[width=0.8\textwidth]{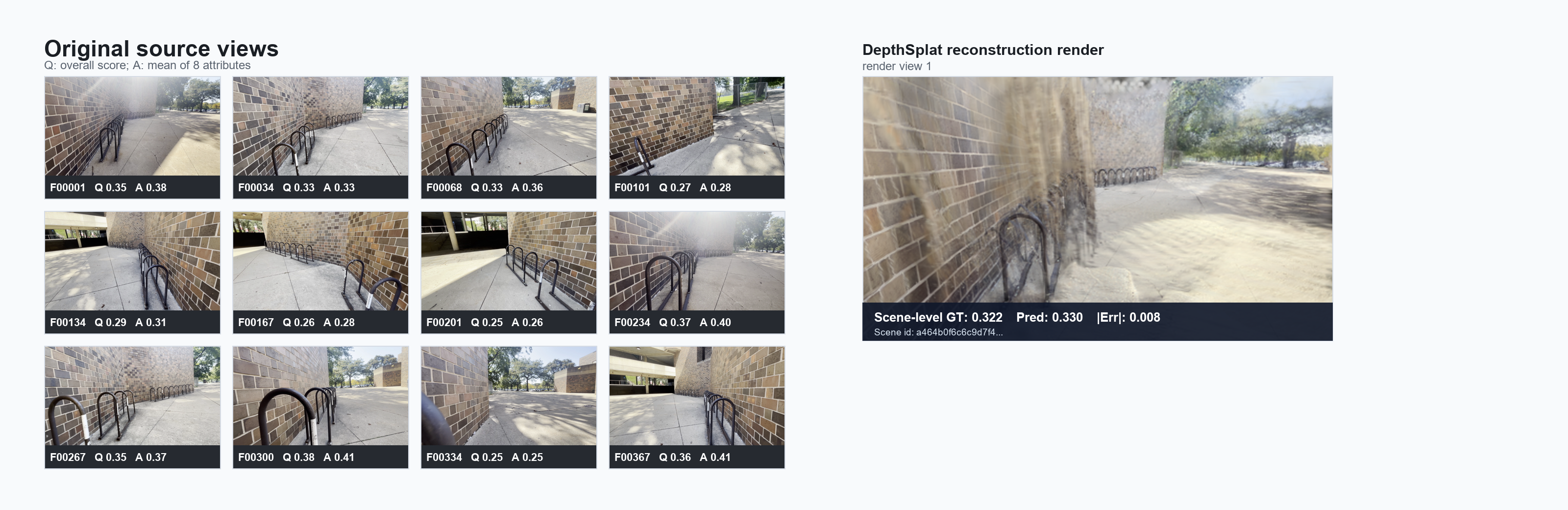}
    \caption{Visualizations of Data Annotation Examples (2).}
    \label{fig:data_examples_vertical_2}
\end{figure*}

\noindent\textbf{Summary.}
Overall, the statistical analysis reveals three key properties of the Aesthetic3D annotations:
(1) a clear cross-dataset distribution shift between DL3DV-10K and Bilarf,
(2) strong inter-attribute correlations indicating that the eight attributes largely capture a shared aesthetic signal rather than independent factors, and
(3) high consistency between attribute-level scores and the overall aesthetic score, particularly under simple aggregation schemes.

These findings support the use of attribute-level scores as an interpretable decomposition of scene aesthetics, while also justifying the use of their unweighted mean as a stable and effective supervision target. At the same time, they highlight the importance of considering dataset bias and attribute redundancy in future model design.

\section{Visualizations of Data Annotation Examples}
\label{appendix:Visualizations of Data Annotation Examples}
Figures~\ref{fig:data_examples_vertical} and~\ref{fig:data_examples_vertical_2} present several visualization examples of the data annotations. The aesthetic annotations of randomly sampled images from different viewpoints are not identical and exhibit noticeable variation. This suggests that, within a 3D scene, evaluating aesthetics solely from individual viewpoints cannot provide a global understanding. In contrast, our annotated global aesthetic score is more representative of the overall aesthetic quality of a 3D scene. In addition, the overall score is very close to, but not exactly the same as, the average score of the eight attributes.

\section{Human Study Validation for Annotation}
\label{appendix:Human Study Validation for Annotation}

We conduct a human study to test whether the ArtiMuse-derived proxy scores (aggregated attribute-level scores) align with human judgments of 3D scenes.

In this analysis, the reference signal, denoted as \emph{Score}, is computed as the equal-weight average of the
eight aesthetic sub-attribute scores for each scene and then rescaled to the same 0--10 range
as the Human Study ratings. This aggregated score corresponds to the proxy supervision signal
used in our framework and serves as the human reference.

\noindent\textbf{Human-study Protocol.}
We organized the rating study with five volunteers. All participants reported
regular experience using mobile phones or cameras for photography or video
creation, and some also reported experience with image retouching or color
adjustment software. The participants therefore had practical exposure to
visual composition, color, lighting, and scene presentation.

Before the formal rating session, each participant was instructed to evaluate
the overall aesthetic quality of a 3D scene in terms of visual appeal and human
preference, rather than focusing solely on geometric accuracy or reconstruction
fidelity. This instruction ensures consistency with the definition of aesthetic
quality used in our annotation pipeline.

\begin{figure*}[t]
    \centering
    \includegraphics[width=0.8\textwidth]{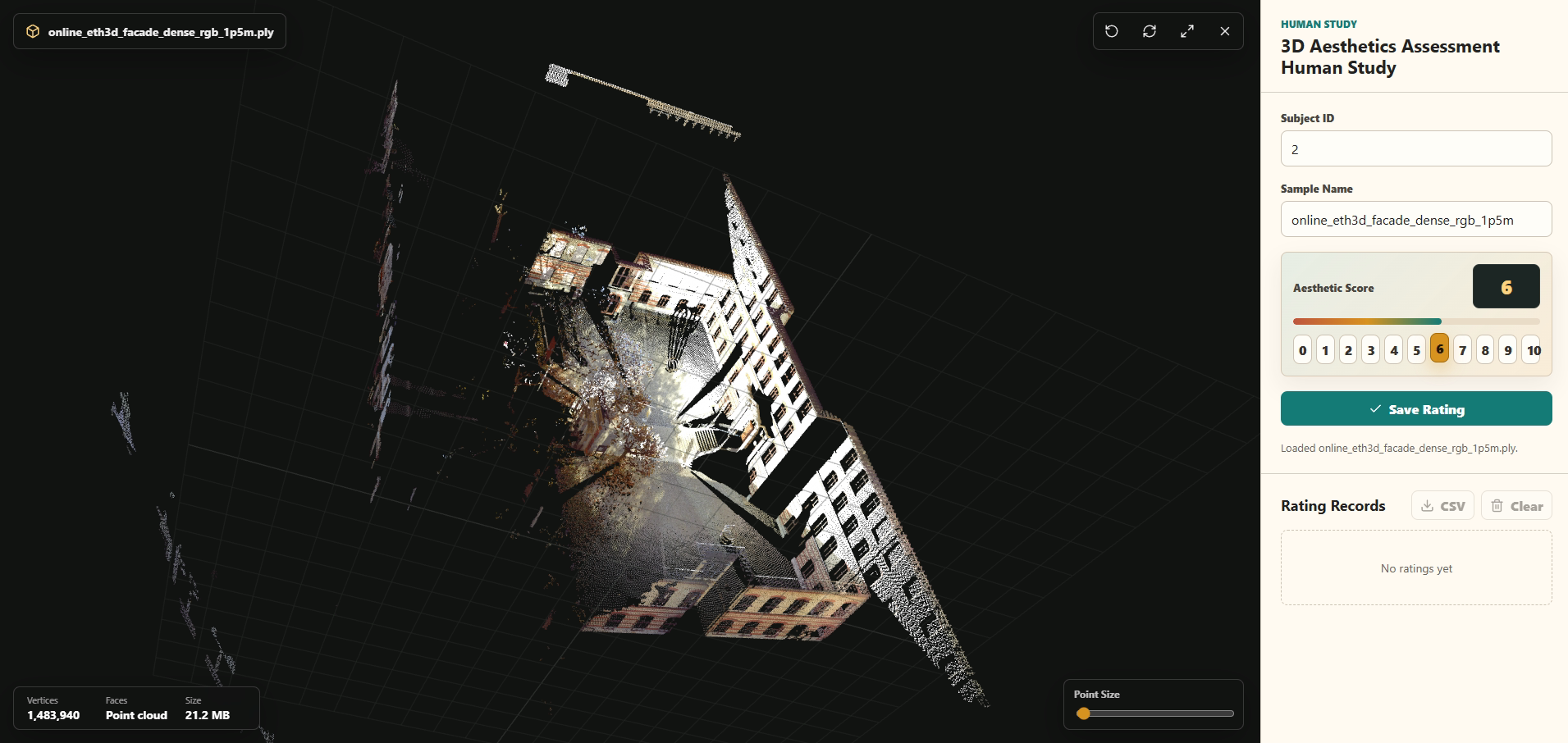}
    \vspace{2pt}
    \includegraphics[width=0.8\textwidth]{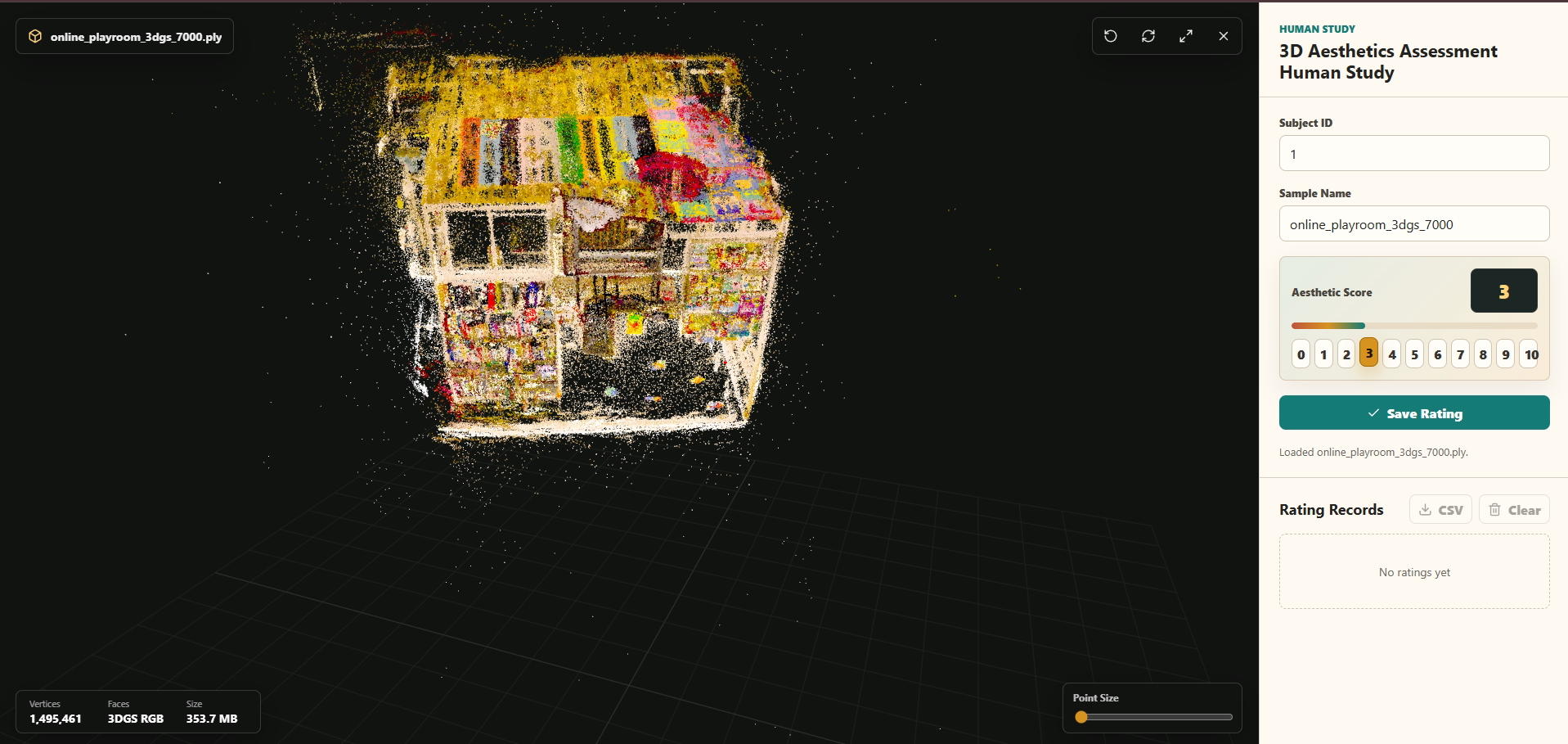}
    \caption{
        Screenshots of the custom rating interface used in the human study.
        The interface provides participants with an interactive 3D scene viewer
        supporting rotation, panning, and zooming, along with an 11-level
        integer rating scale from 0 to 10 for assessing the overall aesthetic
        quality of each scene.
    }
    \label{fig:human_study_interface}
\end{figure*}

Participants used a custom rating interface (refer to Figure~\ref{fig:human_study_interface}) to inspect the 278 3DGS scenes. For each
scene, the interface provided an interactive 3D viewer and an 11-level integer
rating scale from 0 to 10. Participants were allowed to rotate, pan, and zoom
the scene before assigning a score. The average inspection time was approximately
20 seconds per scene, which allows sufficient observation of both global structure
and local details while keeping the overall study feasible. A 10--20 minute break
was provided at approximately 50\% progress to reduce fatigue. Each session lasted
around two hours, and participants were compensated above the local minimum wage.

\begin{figure*}[t]
\centering
\includegraphics[width=0.8\textwidth]{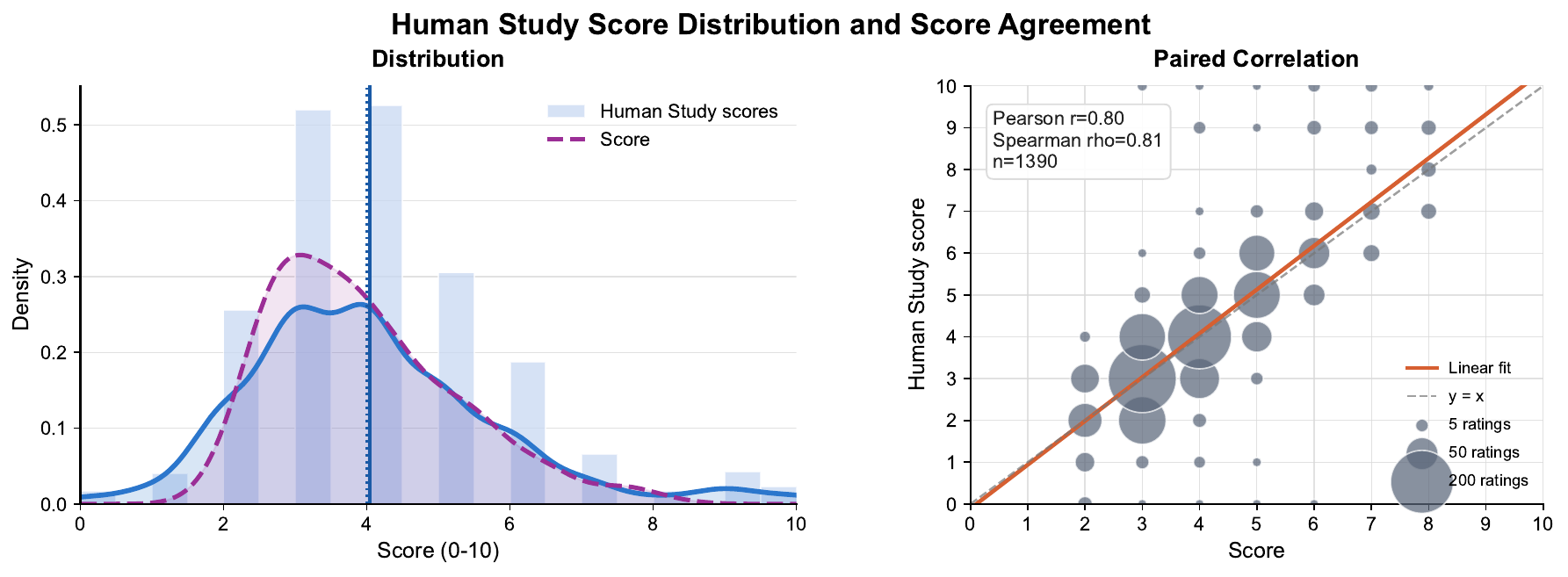}
\caption{
\textbf{Alignment between Human Study ratings and proxy aesthetic scores.}
Left: comparison of the marginal distributions of Human Study ratings and the proxy Score.
Right: paired agreement between integer Score grades and integer Human Study ratings, where
bubble size indicates the number of participant-scene ratings at each pair.
Most ratings concentrate near the diagonal in the mid-score range, indicating strong agreement,
while small off-diagonal bubbles reflect subjective deviations.
}
\label{fig:human_study_score_distribution_correlation}
\end{figure*}

Figure~\ref{fig:human_study_score_distribution_correlation} summarizes the agreement
between Human Study ratings and the proxy Score. The left panel compares the marginal
distributions of the two signals. Referring to Table \ref{tab:human_study_score_reliability}, they exhibit highly similar central tendencies: Human
Study ratings have a mean of 4.04 and a median of 4.00, while Score has a mean of 3.94
and a median of 3.67. Both distributions are positively skewed (Human Study: 0.92,
Score: 0.88), indicating a dense mid-score region with a smaller high-quality tail.

The right panel visualizes paired agreement. Since both axes are discretized into integer
levels, the plot explicitly reflects the discrete nature of human annotation. Most ratings
form large bubbles near the diagonal, indicating that scenes with similar proxy scores are
also judged similarly by human participants. A limited number of off-diagonal bubbles appear
at the extremes, corresponding to subjective outlier responses rather than systematic disagreement.

Beyond distributional similarity, the quantitative agreement is also strong. Using all
1,390 participant-scene ratings, Human Study scores achieve Pearson correlation
$r=0.80$, Spearman correlation $\rho=0.81$, and Kendall correlation $\tau_b=0.603$ with the proxy Score. Agreement among the five raters is Kendall's $W=0.673$. These results indicate
that the proxy aesthetic scores derived from the ArtiMuse-based annotation pipeline are
well aligned with human judgments, both in terms of absolute values and relative ranking.

\begin{table*}[t]
\centering
\caption{
Summary statistics for the pooled DL3DV-10K and BiLaRF reliability analysis.
Human Study ratings are compared with the proxy Score derived from attribute aggregation.
}
\label{tab:human_study_score_reliability}
\begin{tabular}{lrrrrrr}
\toprule
Signal & $n$ & Mean & Median & Std. & Skewness & Range \\
\midrule
\rowcolor{blockB}
Score & 278 & 3.94 & 3.67 & 1.33 & 0.88 & 1.80--8.25 \\
\rowcolor{blockC}
Human Study & 1390 & 4.04 & 4.00 & 1.73 & 0.92 & 0.00--10.00 \\
\midrule
\multicolumn{7}{l}{Pearson $r=0.80$, Spearman $\rho=0.81$, Kendall $\tau_b=0.603$, and inter-rater Kendall $W=0.673$.} \\
\bottomrule
\end{tabular}
\end{table*}

\noindent\textbf{Conclusion.}
Overall, the human study provides strong evidence that the proxy aesthetic scores used in
Aesthetic3D align with human ratings. The agreement in both distribution and ranking supports the
ArtiMuse-based annotation pipeline for scene-level aesthetic assessment.
While human ratings remain discrete and occasionally noisy, their consistency with the proxy
signal supports the use of the proposed annotation strategy as a scalable alternative to
large-scale human labeling.
\section{Supplementary Explanation for Projection-based View Tokenization}
\label{appendix: Supplementary Explanation for Projection-based View Tokenization}
\subsection{Candidate Camera Construction and Sampling}
\label{app:camera}

Each 3D scene is represented by 2048 Gaussian primitives selected via farthest point sampling (FPS).
We normalize the Gaussian centers by subtracting the scene centroid and dividing by the 95th percentile
of the radius distribution, ensuring a consistent scale across scenes.
FPS improves spatial coverage but does not preserve Gaussian density or target reconstruction artifacts. Fine structures can therefore be missed, and the current input does not encode scale or rotation even when an anisotropic Gaussian is selected.

The candidate views used in our model are not synthetically generated. Instead, they are directly selected
from the set of real cameras provided by the source datasets, all of which come with calibrated intrinsic
and extrinsic parameters.

For each scene, we consider all available source cameras and perform uniform binning over the spherical
azimuth--elevation space, where each camera is represented by the unit direction vector from the normalized
scene origin to the camera center. We then select 32 cameras by uniformly sampling across these bins.
If a scene contains fewer than 32 cameras, all available cameras are used.

Each selected candidate view retains its original camera parameters, including the intrinsic matrix
$K_v$ and the extrinsic matrix $E_v = [R_v \,|\, t_v]$. We do not synthesize new camera poses, do not
assume shared focal lengths, and do not enforce any canonical camera configuration (e.g., look-at or
fixed up-vector). The only transformation applied is the same normalization used for the Gaussian
primitives, which is consistently applied to the camera centers.

These candidate views are used purely as probe cameras for geometric projection within the model,
and no RGB rendering is performed. The same camera construction procedure is used during training and evaluation. We reinitialize the sampling seed each training epoch and use a fixed seed for evaluation.

We adopt real source cameras instead of synthetically generated spherical cameras for three reasons.
First, 3D Gaussian Splatting is only reliable near observed viewpoints, and synthetic cameras may fall
into unseen regions where Gaussian support is sparse, leading to empty projections and noisy features.
Second, the per-view aesthetic supervision (from ArtiMuse) is defined on the original images. Using the
same cameras ensures geometric consistency between input representations and supervision signals.
Third, the intrinsic parameters of source cameras naturally reflect real-world variations in imaging
devices (e.g., wide-angle or telephoto lenses), whereas enforcing shared intrinsics would distort the
projected spatial structure.

By combining real camera selection with spherical binning, our strategy achieves both angular coverage
and distribution alignment with the original training data.

\subsection{Projection-based View Tokenization - Patch Aggregation and View-Conditioned Context}
\label{appendix:Projection-based View Tokenization}
To obtain view-level evidence without rendering RGB images, we project the 3D scene tokens into candidate views. This design is motivated by 3D-to-multi-view representation learning~\citep{chen2025point}, where camera geometry is used to derive multi-view evidence from 3D point tokens.

For the $v$-th candidate view, each 3D point is projected onto the view plane using the corresponding intrinsic matrix $\mathbf{K}_v$, expressed in normalized image coordinates, and world-to-camera extrinsic matrix $\mathbf{E}_v=[\mathbf{R}_v \mid \mathbf{t}_v]$:
\begin{equation}
    \mathbf{u}_{iv}
    =
    \mathbf{K}_v
    \Pi
    \left(
        \mathbf{E}_v
        \tilde{\mathbf{p}}_i
    \right),
\end{equation}
where $\tilde{\mathbf{p}}_i=[\mathbf{p}_i^\top,1]^\top$ is the homogeneous form of $\mathbf{p}_i$, and $\Pi([x,y,z]^\top)=[x/z,\;y/z,\;1]^\top$ denotes perspective division to normalized image coordinates.

The projected view plane is discretized into a $G \times G$ patch grid. Let $\mathcal{T}_{v,m}$ be the set of scene tokens projected into patch $m$ of view $v$. For each projected patch cell, we first aggregate the projected scene tokens as:
\begin{equation}
    \mathbf{b}_{v,m}
    =
    \begin{cases}
    \mathrm{Mean}(\mathcal{T}_{v,m})
    +
    \mathrm{Max}(\mathcal{T}_{v,m}),
    &
    \mathcal{T}_{v,m}\neq\emptyset,\\
    \mathbf{e}_{\emptyset},
    &
    \mathcal{T}_{v,m}=\emptyset,
    \end{cases}
\end{equation}
where $\mathbf{e}_{\emptyset}\in\mathbb{R}^{D}$ is a learned empty-cell token.

The patch token is then conditioned on the geometry of view $v$:
\begin{equation}
    \mathbf{r}_{v,m}
    =
    \phi_{\mathrm{cell}}(\mathbf{b}_{v,m})
    +
    \mathbf{a}_v ,
\end{equation}
where $\phi_{\mathrm{cell}}$ maps the pooled projected token to $\mathbb{R}^{D}$.

The view-conditioned context $\mathbf{a}_v\in\mathbb{R}^{D}$ is computed as:
\begin{equation}
\begin{aligned}
\mathbf{a}_v = \phi_{\mathrm{geom}}\!\Bigl(\bigl[&
    \mathbf{R}_v^{\!\top}\mathbf{e}_x;\,
    \mathbf{R}_v^{\!\top}\mathbf{e}_y;\\
&\mathbf{R}_v^{\!\top}\mathbf{e}_z;\,
    \mathbf{o}_v;\,
    \log(1+\tilde f_x);\\
&\log(1+\tilde f_y);\,
    \tilde c_x;\,
    \tilde c_y\bigr]\Bigr).
\end{aligned}
\end{equation}
where $\mathbf{e}_x,\mathbf{e}_y,\mathbf{e}_z$ are the canonical basis vectors, $\mathbf{o}_v=-\mathbf{R}_v^{\!\top}\mathbf{t}_v$ is the camera center in the scene coordinate system, and $\tilde f_x,\tilde f_y,\tilde c_x,\tilde c_y$ are normalized intrinsic parameters derived from $\mathbf{K}_v$.

The resulting patch tokens are refined by a lightweight view Transformer and summarized into a single view token descriptor $\mathbf{v}_v$. In this way, each descriptor is constructed from projected 3D scene tokens conditioned on view geometry, rather than from rendered image features or a pretrained 2D backbone.

\section{Full Implementation Details of Aes3DGSNet}
\label{app:aes3dgsnet_details}

This section provides additional implementation details for the proposed Aes3DGSNet. The model components, notation, and training objective follow Section~4, including Gaussian scene encoding, projection-based view tokenization, candidate view selection, multi-view fusion and regression, and the projection, view pooling, top-\ensuremath{K} selection, fusion, and pairwise hinge ranking losses defined in Eq.~(3)--(8).

\subsection{Computational Resources}
Aes3DGSNet training takes approximately 40 minutes on a single NVIDIA GeForce RTX 5090 GPU with 32 GB VRAM. ArtiMuse annotation for 278 3D scenes in Aesthetic3D was performed on 2 NVIDIA H100 GPUs and required approximately 30 GPU-hours.

\subsection{Aes3DGSNet Architecture}
\label{app:aes3dgsnet_implementation}

Aes3DGSNet directly takes 3DGS primitives as input, rather than using rendered RGB images as direct input. For each 3DGS scene, we sample \ensuremath{N{=}2048} Gaussian primitives using farthest point sampling (FPS). The input feature of each primitive follows Section~4.1 and consists of the normalized Gaussian center, RGB attribute, and the unit direction from the scene center to the Gaussian center.

\noindent\textbf{Gaussian Scene Encoding.}
We use a 4-block point Transformer encoder for Gaussian scene encoding. Unless otherwise specified, all Transformer modules use hidden dimension \ensuremath{D{=}192}, 4 attention heads, MLP ratio 2, and dropout 0.1. The encoder outputs point-level scene tokens, and a single-query attention pooling module further produces a scene-global token. This scene-global token is used in the candidate view selection stage to provide scene-level context for the view selector. For scenes with fewer valid sampled primitives, padding masks are used throughout the point encoder and attention pooling.

\noindent\textbf{Projection-based View Tokenization.}
We use \ensuremath{V{=}32} candidate views for projection-based view tokenization. These 32 candidate views are internal views used by Aes3DGSNet for geometric projection, view tokenization, and view selection. They are not the same as the rendered RGB views used by rendered-view baselines. The rendered-view baselines instead use all available rendered views from the precomputed rendered-view table, and the number of rendered views can vary across scenes.

For each candidate view, Gaussian scene tokens are projected onto the view plane following Eq.~(3), and the projected plane is discretized into a \ensuremath{14{\times}14} patch grid. Tokens falling into the same patch cell are aggregated by the mean-max view pooling operation in Eq.~(4). Empty patch cells are masked in the subsequent Transformer. The resulting patch tokens are refined by a 2-block lightweight view Transformer and pooled into one view descriptor. We inject view-geometry/pose encoding to make the descriptor aware of camera geometry.

\noindent\textbf{Candidate View Selection.}
The candidate view selector is implemented as a 2-block listwise Transformer. Its input sequence contains the candidate view tokens, the scene-global token, and two learnable control tokens. The view selector predicts one utility logit for each valid candidate view and computes sparse top-\ensuremath{K} selection weights following Eq.~(5). We set the temperature to \ensuremath{\tau{=}1.0} and select \ensuremath{K{=}8} views. In implementation, hard top-\ensuremath{K} selection is performed on detached utility logits, and the softmax weights of the selected views are renormalized to obtain sparse view weights. Gradients flow through the selected-view soft weights.

\noindent\textbf{Multi-view Fusion and Scene-level Regression.}
Multi-view fusion and scene-level regression follow Eq.~(6). The selected view descriptors are aggregated by sparse weighted summation to form a multi-view scene representation. This representation is then passed through a residual MLP, LayerNorm, and a 3-layer scalar regressor to predict the scene-level aesthetic score. In the final configuration, the regression head does not directly re-consume the scene-global token. Instead, prediction uses aggregated selected-view evidence.

\noindent\textbf{Supervision Targets and Ablation Variants.}
The two main variants, Aes3DGSNet (total target) and Aes3DGSNet (8-attr mean target), use the same architecture, optimizer, and training settings, differing only in the supervision target. The former uses the scene-level target aggregated from the ArtiMuse overall score, while the latter uses the scene-level target aggregated from the mean of the eight attribute-level scores. The ablation variants in Table~2 modify specific modules, such as removing the scene-global token, removing learnable control tokens, replacing learned selection with uniform aggregation, changing \ensuremath{K}, or removing geometric projection.

\subsection{Training Objective and Optimization}
\label{app:training_objective}

Aes3DGSNet is trained for 16 epochs. We use AdamW with learning rate \ensuremath{5{\times}10^{-5}}, weight decay \ensuremath{10^{-4}}, and batch size 4. Training uses cosine learning-rate decay, bfloat16 mixed precision, and gradient clipping with norm 1.0.

The objective follows Section~4.5 and combines Huber regression loss with a pairwise hinge ranking loss:
\begin{equation}
    \mathcal{L}
=
\mathrm{Huber}_{\delta=1}(\hat{y}, y)
+
\lambda_{\mathrm{rank}}\mathcal{L}_{\mathrm{rank}} .
\end{equation}
The Huber loss uses \ensuremath{\delta{=}1}. The ranking term follows the hinge formulation in Eq.~(8), i.e., the \ensuremath{\max(0,\cdot)} form, rather than a soft-hinge or softplus formulation. We set \ensuremath{\lambda_{\mathrm{rank}}{=}0.1}, the ranking margin to \ensuremath{m{=}0.05}, and include a pair in the ranking loss only when the absolute target-score gap is larger than \ensuremath{\epsilon{=}0.03}. This avoids strong ordinal constraints on pairs with small, potentially unstable target differences.

\section{Full Implementation Details of Baselines}
\label{app:baseline_details}

This section describes the implementation details of all baseline methods.

\subsection{Train-split Linear Calibration}
\label{app:linear_calibration}

For the four zero-shot 2D image aesthetic baselines (ArtiMuse, AADB, NIMA, and TANet), the numerical scale of raw predictions may not be fully aligned with the scene-level regression target. Therefore, we apply the same train-split linear calibration to all four baselines. Specifically, we fit a linear mapping on the training split:
\begin{equation}
\tilde{y} = a\hat{y} + b ,
\end{equation}
where \ensuremath{\hat{y}} is the raw scene-level prediction from the baseline, and \ensuremath{\tilde{y}} is the calibrated prediction. The parameters \ensuremath{a} and \ensuremath{b} are fitted using only predictions and labels from the training split. Once fitted, they are fixed and directly applied to holdout/test scenes.

Test-set labels are reserved for final evaluation.

\subsection{ArtiMuse Zero-shot Baseline}
\label{app:artimuse_baseline}

The ArtiMuse zero-shot baseline directly applies the image aesthetic assessment model to rendered RGB views, without fine-tuning on Aesthetic3D. For each scene, ArtiMuse first predicts view-level aesthetic scores for all available rendered views. These scores are then averaged to obtain a scene-level prediction, which is further mapped to the scene-level target scale through train-split linear calibration.

This baseline addresses a direct question: whether applying a strong 2D image aesthetic prior to 3DGS rendered views, followed by simple scene-level aggregation, is already sufficient to recover the scene-level proxy aesthetic target constructed in this work. This baseline does not include 3DGS-aware primitive modeling, projection-based view tokenization, or learned candidate view selection.

\subsection{Reproduced Quality Assessment Baselines}
\label{app:qa_baselines}

To further compare Aes3DGSNet with existing no-reference 3D quality assessment methods, we reproduce five additional baselines and group them into two categories: point-cloud-based quality assessment and 3D Gaussian Splatting (3DGS)-specific quality assessment.

All reproduced baselines are adapted to the same Aesthetic3D benchmark protocol, including the unified data manifest, train/test splits, and proxy aesthetic labels. All methods are evaluated on the normalized \ensuremath{[0,1]} scale for MAE and RMSE. PLCC is computed after nonlinear logistic fitting, while SRCC and KRCC are computed on raw predictions. Results are averaged over seeds \ensuremath{\{7,13,42\}}.

\noindent\textbf{Point-based Quality Assessment Baselines}\par

\noindent\textbf{MM-PCQA \cite{baseline1}.}
MM-PCQA is a multi-modal no-reference point-cloud quality assessment method that jointly leverages colored point clouds, 2D projections, and local geometry patches. Since our data consists of 3D Gaussian reconstructions, we convert each scene into a normalized \texttt{xyzrgb} point cloud, where RGB values are recovered from SH DC coefficients.

For each scene, we generate four orthographic projection images and six local geometry patches (2048 points each). These inputs are fed into the image branch, point-cloud branch, and cross-modal fusion module. The image branch uses \ensuremath{224{\times}224} crops with ImageNet normalization.

We retrain the model on our benchmark split using L2RankLoss, Adam optimizer (lr=\ensuremath{5{\times}10^{-5}}, wd=\ensuremath{10^{-4}}), batch size 8, StepLR scheduler (step size 8, gamma 0.9), for 50 epochs.

\noindent\textbf{LRL-GQA \cite{baseline2}.}
LRL-GQA is a geometry-only no-reference quality assessment method for colorless point clouds. We adapt it by using only Gaussian center coordinates as input and discarding all color and appearance attributes.

Each scene is normalized and divided into 64 local geometry patches with 512 points each using FPS anchors and KNN grouping. We retain the original DGCNN-style patch encoder and weighted aggregation module.

The model is trained with SmoothL1 loss and list-wise ranking loss using AdamW (lr=\ensuremath{10^{-4}}, wd=\ensuremath{10^{-4}}), batch size 2, ranking-loss weight 0.2, KNN size \ensuremath{k{=}20}, for 40 epochs.

\noindent\textbf{Stochastic NR-PCQA \cite{baseline3}.}
Stochastic NR-PCQA models the stochastic nature of subjective quality ratings using a CVAE-based framework. To adapt it to Gaussian scenes, we render random RGB-D multi-view projections from Gaussian centers and SH-DC colors.

Each sample consists of four views at \ensuremath{480{\times}480} resolution. We implement the model following the architecture and training protocol described in the paper, as the official repository does not provide a complete training pipeline.

The model uses latent dimension 3, feature channels 32, and a ResNet-50 backbone. Training uses Adam (lr=\ensuremath{2.5{\times}10^{-5}}, betas \ensuremath{(0.5,0.999)}), batch size 8, \ensuremath{\alpha{=}0.4}, for 200 epochs. During inference, we average 37 stochastic prior samples per scene.

\noindent\textbf{3D Gaussian Splatting Quality Assessment Baseline}\par

\noindent\textbf{GSOQA \cite{GSP02}.}
GSOQA is specifically designed for perceptual quality assessment of 3D Gaussian Splatting and directly operates on Gaussian primitives. Since the official code and pretrained models are unavailable, we reimplement the method following the architecture and training protocol described in the paper.

Each Gaussian primitive is represented as a 59D vector including center coordinates, opacity, scale, rotation, and SH features. Available SH DC coefficients are used directly, while missing higher-order SH coefficients are zero-padded.

For each scene, we randomly downsample to 2048 primitives, select 64 FPS centers, and group 32 KNN neighbors per center. The model uses a Gaussian-MAE-style local encoder (hidden dim 384, depth 12, 6 heads), followed by 3 graph-attention refinement blocks with 16 neighbors.

Training uses AdamW (lr=\ensuremath{10^{-4}}, wd=\ensuremath{10^{-4}}), batch size 32, OneCycleLR scheduling, and loss \ensuremath{0.5\mathcal{L}_{\mathrm{lin}} + 0.5\mathcal{L}_{\mathrm{mon}}} for 100 epochs.

\noindent\textbf{MUGSQA-DBCNN \cite{baseline4}.}
We implement a rendered-view no-reference IQA baseline following the MUGSQA paradigm, using DBCNN as the underlying image quality predictor. Since official MUGSQA benchmark weights are not publicly available, we initialize DBCNN with KonIQ-10k pretrained weights from IQA-PyTorch / PyIQA. The model contains 15.3M parameters.

Each rendered view is resized to $512 \times 512$ and independently scored by DBCNN. Scene-level predictions are obtained by mean aggregation over all available rendered views. The rendered-view table contains 278 scenes and 2297 views in total, with an average of 8.26 views per scene, at an estimated cost of 712.2 GFLOPs per scene.

For supervision, we replace the original MOS target in $[0,5]$ with our proxy aesthetic label 8-attr mean score. Each rendered view inherits the scene-level label during training. For evaluation, predictions are mapped back to the raw $[0,1]$ scale using 8-attr mean score for MAE and RMSE computation.

The training protocol follows the same dataset split as other baselines, using stratified holdout splits with seeds $\{13,42,7\}$. The model is trained for 16 epochs using AdamW (learning rate $1\times10^{-4}$, weight decay $1\times10^{-4}$), batch size 8, SmoothL1Loss ($\beta=0.5$), bfloat16 mixed precision, and gradient clipping with norm 1.0. We use an unfrozen backbone and apply random horizontal flip with mild color jitter as augmentation.

This is a MUGSQA-style rendered-view NR-IQA transfer baseline, rather than the official MUGSQA model.

\subsection{Aggregation Rules}
\label{app:aggregation_rules}

Different baselines use different aggregation rules according to their supervision granularity and pipeline design.

The ArtiMuse zero-shot baseline has no trainable view selector and is not fine-tuned on Aesthetic3D. Therefore, we use mean aggregation to average ArtiMuse predictions over all available rendered views into a scene-level prediction.

We use a fixed view selection budget of \ensuremath{K{=}8}.

\subsection{Proxy-label Setting}
\label{app:proxy_label_setting}

The scene-level supervision in Aesthetic3D is derived from an IAA-based annotation pipeline. Therefore, the supervision signal should be understood as IAA-derived proxy scene-level aesthetic labels, rather than direct human ground-truth aesthetic ratings. The goal of this work is to study 3DGS scene-level aesthetic regression under this proxy-label benchmark and to compare direct 3DGS-aware modeling against rendered-view 2D baselines.

ArtiMuse scores the original captured images, rather than views rendered from the reconstructed 3DGS scenes. The proxy labels may therefore miss reconstruction-specific artifacts, such as floaters. The human study in Section~\ref{appendix:Human Study Validation for Annotation} shows alignment with overall human ratings, but does not measure sensitivity to every such artifact.

Under this setting, the ArtiMuse zero-shot baseline has a special but meaningful role. Since the benchmark labels are themselves constructed from an IAA pipeline, this baseline tests whether directly applying a similar 2D aesthetic prior to rendered RGB views, followed by simple aggregation, is already sufficient to recover the scene-level proxy target. If this baseline remains limited, it suggests that rendered-view 2D scoring alone is insufficient to fully model the 3DGS scene-level aesthetic signal defined in this work. This further supports the need for direct 3DGS-aware multi-view aesthetic regression from Gaussian primitives.

\section{Full Ablation Studies}
\label{appendix:Full Ablation Studies}

The supplementary ablations further clarify where the gains of Aes3DGSNet come from and show that the full design is consistently the strongest across all five metrics in Table~\ref{tab:appendix_ablation_extra}. Our full model achieves the best overall performance, with PLCC/SRCC/KRCC of $\mathbf{.567 \pm .036}$/$\mathbf{.554 \pm .043}$/$\mathbf{.389 \pm .032}$ and the lowest MAE/RMSE of $\mathbf{.084 \pm .008}$/$\mathbf{.110 \pm .005}$, indicating that the proposed components contribute in a complementary manner rather than through a single dominant factor.

First, the projection-based view tokenization ablations show that high-quality view descriptors require both spatially structured projection and adaptive patch-level aggregation. Removing the patch transformer causes a clear drop to $.514 \pm .028$ PLCC and $.512 \pm .022$ SRCC, confirming that contextual interaction among projected patches is important. Replacing the default projection setup with a coarser $7{\times}7$ projection grid also reduces rank and correlation performance, suggesting that finer spatial discretization better preserves appearance cues relevant to aesthetic judgment. Likewise, replacing the mean+max scatter with mean-only scatter leads to weaker results, showing that retaining both average and saliency-like responses is beneficial. Among these variants, mean patch pooling performs the worst, with PLCC/SRCC/KRCC falling to $.492 \pm .044$/$ .475 \pm .045$/$ .326 \pm .032$, which highlights that query-based adaptive pooling is substantially more effective than uniform averaging for summarizing projected patch features.

Second, the Gaussian input representation ablations confirm that the model benefits from a compact but informative attribute design. Using xyz only leads to the largest performance degradation in the table, dropping to $.393 \pm .020$ PLCC and $.413 \pm .035$ SRCC, while also increasing MAE/RMSE to $.101 \pm .009$/$ .126 \pm .002$. This indicates that geometry alone is insufficient for aesthetic quality prediction. Adding RGB already recovers most of the lost performance, reaching $.555 \pm .016$ PLCC and $.526 \pm .038$ SRCC, which shows that appearance cues are essential. Interestingly, using xyz plus all raw Gaussian attributes does not outperform the compact representation used in the full model, and remains below it on all metrics. This suggests that simply exposing more low-level Gaussian parameters is not necessarily helpful, whereas the proposed compact xyz+rgb+direction representation strikes a better balance between informativeness and learnability.

Third, the candidate view selection and probe robustness experiments show that the gains are not solely due to candidate-view coverage, but also to how views are weighted and sampled. The selected-uniform weighting variant achieves $.525 \pm .027$ PLCC and $.533 \pm .021$ SRCC, which remains competitive but still below the full model, indicating that the selector contributes not only through top-K view selection but also through its non-uniform learned weighting of selected views. Replacing structured probe generation with random probe sampling yields $.534 \pm .037$ PLCC and $.519 \pm .027$ SRCC, again trailing the full model. This result suggests that Aes3DGSNet is reasonably robust to the precise choice of candidate views, while still benefiting from the structured coverage provided by selecting real source cameras using uniform azimuth--elevation bins.

Overall, these ablations support three main conclusions. (1) Projection-based view tokenization should preserve local spatial structure and use adaptive aggregation rather than simple averaging. (2) Aesthetic prediction requires both geometry and appearance, but a compact, task-oriented Gaussian representation is preferable to naively using all available attributes. (3) Performance gains further rely on learned view weighting and structured probe coverage, both of which contribute beyond the base effect of multi-view sampling.

\begin{table*}[t]
\centering
\small
\caption{Additional ablations of Aes3DGSNet in the appendix. Results are mean $\pm$ std over three seeds $\{7,13,42\}$. Best results are \textbf{bold}.}
\label{tab:appendix_ablation_extra}
\resizebox{\textwidth}{!}{%
\begin{tabular}{lccccc}
\toprule
Variant & PLCC $\uparrow$ & SRCC $\uparrow$ & KRCC $\uparrow$ & MAE $\downarrow$ & RMSE $\downarrow$ \\
\midrule
\rowcolor{blockB}
\multicolumn{6}{l}{\textit{\textbf{Full model}}}\\
\rowcolor{blockB}
Aes3DGSNet (ours) & $\mathbf{.567 \pm .036}$ & $\mathbf{.554 \pm .043}$ & $\mathbf{.389 \pm .032}$ & $\mathbf{.084 \pm .008}$ & $\mathbf{.110 \pm .005}$ \\

\addlinespace[2pt]
\rowcolor{blockC}
\multicolumn{6}{l}{\textit{\textbf{(1) Projection-based view tokenization}}}\\
\rowcolor{blockC}
w/o patch transformer & $.514 \pm .028$ & $.512 \pm .022$ & $.359 \pm .020$ & $.089 \pm .007$ & $.114 \pm .004$ \\
\rowcolor{blockC}
$7{\times}7$ projection grid & $.533 \pm .029$ & $.513 \pm .051$ & $.356 \pm .041$ & $.088 \pm .006$ & $.110 \pm .010$ \\
\rowcolor{blockC}
mean-only scatter & $.541 \pm .024$ & $.517 \pm .054$ & $.360 \pm .040$ & $.088 \pm .003$ & $.114 \pm .002$ \\
\rowcolor{blockC}
mean patch pooling & $.492 \pm .044$ & $.475 \pm .045$ & $.326 \pm .032$ & $.089 \pm .005$ & $.113 \pm .008$ \\

\addlinespace[2pt]
\rowcolor{blockD}
\multicolumn{6}{l}{\textit{\textbf{(2) Gaussian input representation}}}\\
\rowcolor{blockD}
xyz only & $.393 \pm .020$ & $.413 \pm .035$ & $.284 \pm .029$ & $.101 \pm .009$ & $.126 \pm .002$ \\
\rowcolor{blockD}
xyz + rgb & $.555 \pm .016$ & $.526 \pm .038$ & $.369 \pm .028$ & $.088 \pm .005$ & $.113 \pm .003$ \\
\rowcolor{blockD}
xyz + full Gaussian attrs & $.511 \pm .023$ & $.527 \pm .034$ & $.366 \pm .031$ & $.090 \pm .008$ & $.116 \pm .009$ \\

\addlinespace[2pt]
\rowcolor{blockE}
\multicolumn{6}{l}{\textit{\textbf{(3) Candidate view selection and probe robustness}}}\\
\rowcolor{blockE}
selected-uniform weighting & $.525 \pm .027$ & $.533 \pm .021$ & $.372 \pm .004$ & $.087 \pm .003$ & $.112 \pm .005$ \\
\rowcolor{blockE}
random probe sampling & $.534 \pm .037$ & $.519 \pm .027$ & $.364 \pm .020$ & $.088 \pm .002$ & $.115 \pm .003$ \\

\bottomrule
\end{tabular}%
}
\end{table*}

\section{Sensitivity to Reconstruction Views}
\label{app:reconstruction_view_sensitivity}
We tested how the number and placement of reconstruction views affect predictions on 12 held-out scenes. Each scene was reconstructed with 32 views spread along the camera trajectory, 10 similarly spread views, and three separate one-sided sets of 10 views from the beginning, middle, and end of the trajectory. The reconstruction pipeline and Aes3DGSNet checkpoint were fixed. Table~\ref{tab:reconstruction_view_sensitivity} compares predicted scores with the 32-view reference.

\begin{table}[t]
\centering
\small
\caption{Effect of reconstruction views on predicted scores for 12 held-out scenes. Absolute and signed differences use the 32-view reconstruction as reference. Spearman $\rho$ compares scores across scenes.}
\label{tab:reconstruction_view_sensitivity}
\resizebox{\linewidth}{!}{%
\begin{tabular}{lcccc}
\toprule
Reconstruction views & Mean abs. & Median abs. & Mean signed & $\rho$ \\
\midrule
\rowcolor{blockB}
32, uniform & -- & -- & -- & 1.000 \\
\rowcolor{blockC}
10, uniform & .152 & .121 & +.060 & .811 \\
\rowcolor{blockD}
10, one-sided & .206 & .181 & +.172 & .916 \\
\bottomrule
\end{tabular}%
}
\end{table}

For the one-sided setting, the reported $\rho$ uses the average of the three directional scores for each scene. It should not be read as the rank correlation of a single one-sided reconstruction. Across the three directions, the mean within-scene standard deviation is .131 and the mean score range is .250 (maximum .602). Thus, incomplete capture can change the predicted aesthetic score, partly because it changes the reconstructed asset being assessed.

\section{Analysis of Trivial Predictors}
\label{app:trivial_predictor}

\begin{table}[t]
\centering
\caption{Comparison with trivial predictors.}
\label{tab:trivial_predictor}
\setlength{\tabcolsep}{4pt}
\begin{tabular}{lcc}
\toprule
Method & MAE $\downarrow$ & RMSE $\downarrow$ \\
\midrule
Mean predictor & $0.108 \pm 0.009$ & $0.128 \pm 0.008$ \\
Median predictor & $0.111 \pm 0.011$ & $0.134 \pm 0.014$ \\
\midrule
Aes3DGSNet (ours) & $\mathbf{0.084 \pm 0.008}$ & $\mathbf{0.110 \pm 0.005}$ \\
\bottomrule
\end{tabular}
\end{table}
Given the relatively small standard deviation of the scene-level aesthetic scores in Aesthetic3D (approximately $0.13$), a potential concern is that a trivial constant predictor (e.g., predicting the training-set mean) may achieve competitive absolute-error metrics such as RMSE. To explicitly examine this, we evaluate both a mean predictor, defined as $\hat{y} = \frac{1}{|\mathcal{D}_{\text{train}}|} \sum_{i \in \mathcal{D}_{\text{train}}} y_i$, and a median predictor. As shown in Table~\ref{tab:trivial_predictor}, these predictors indeed obtain non-trivial MAE and RMSE values due to the concentrated score distribution, with RMSE close to the dataset standard deviation as expected. However, Aes3DGSNet consistently achieves lower MAE and RMSE than these trivial baselines, indicating that the model learns a meaningful mapping beyond constant prediction. More importantly, trivial predictors produce degenerate correlation metrics (PLCC, SRCC, KRCC), as they assign identical scores to all scenes and thus cannot capture any relative ordering. In contrast, Aes3DGSNet achieves strong rank correlations, demonstrating its ability to model relative aesthetic differences across scenes. This highlights that while absolute-error metrics can be influenced by the limited variance of the dataset, ranking-based metrics remain essential for evaluating aesthetic assessment, where relative preference is the primary objective.

\begin{figure*}[t]
    \centering
    \includegraphics[width=0.8\textwidth]{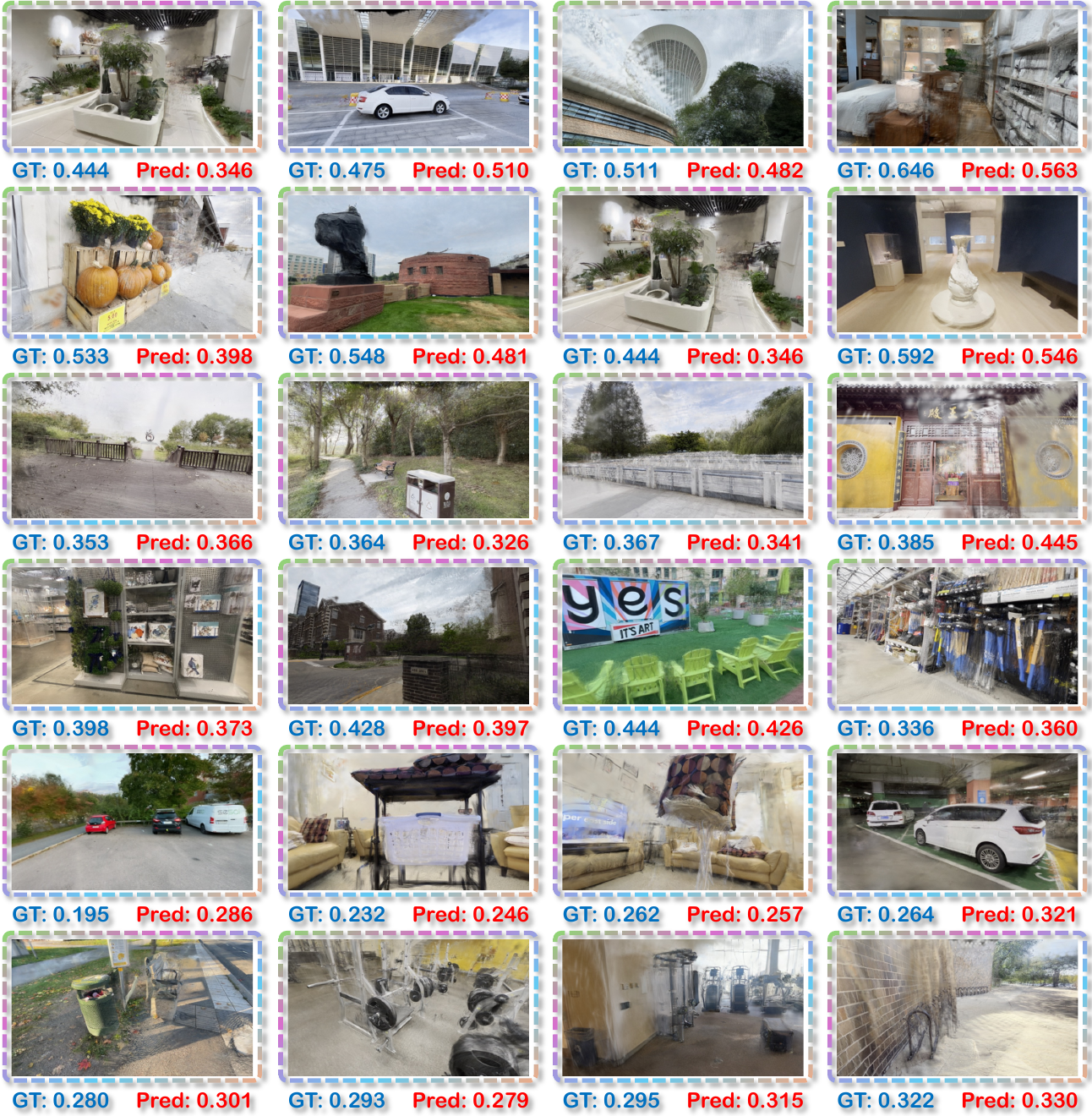}
    \caption{Examples of Aes3DGSNet assessments of 3DGS scenes. "GT" denotes the proxy target, and "Pred" denotes the score predicted by Aes3DGSNet.}
    \label{fig:Visualizations_Aes3DGSNet}
\end{figure*}
\section{Visualizations of Assessment Examples}
\label{appendix:Visualizations}

Figure~\ref{fig:Visualizations_Aes3DGSNet} presents several qualitative examples of aesthetic assessment results produced by Aes3DGSNet on 3DGS scenes. For each example, we show the proxy target (GT) and the predicted score (Pred). Overall, the predicted scores are generally close to the proxy targets, indicating that Aes3DGSNet is able to capture the major aesthetic characteristics of diverse 3D scenes.

From the visualization results, we observe that scenes with cleaner composition, better lighting conditions, clearer structural organization, and more visually pleasing content tend to receive relatively higher scores. In contrast, scenes containing cluttered layouts, poor illumination, motion blur, weak structure, or less attractive contents are usually assigned lower scores. This suggests that the proposed model is able to learn meaningful aesthetic cues that are consistent with the proxy annotations.

Meanwhile, some discrepancies between prediction and proxy target can also be observed. In a few cases, the model tends to underestimate scenes with moderately good aesthetic quality, while in other cases it slightly overestimates scenes whose visual quality is affected by blur, occlusion, or complex background details. These errors are reasonable given the subjective nature of aesthetic evaluation and the large diversity of scene appearance. In particular, for 3D scene aesthetics, the overall perception is often influenced by multiple factors, including composition, color harmony, content richness, depth structure, and viewpoint-dependent visual quality, making precise prediction challenging.

Nevertheless, the qualitative examples demonstrate that Aes3DGSNet produces predictions with reasonable trends across a wide range of scenes, from low-quality everyday environments to more visually appealing indoor and outdoor spaces. These results further verify the effectiveness of the proposed method for aesthetic assessment of 3DGS scenes.

\subsection{Same-scene Comparison Across Methods}
Table~\ref{tab:same_scene_method_comparison} compares predictions for three held-out scenes. We chose one scene from each low-, medium-, and high-score region based only on its proxy target, without using method errors for selection. Figure~\ref{fig:Visualizations_Aes3DGSNet} shows our scores and proxy targets. This table compares four methods.

\begin{table}[!h]
\centering
\small
\caption{Predictions for the same three scenes across methods. The target is the scene-level mean of eight proxy aesthetic attributes.}
\label{tab:same_scene_method_comparison}
\setlength{\tabcolsep}{2pt}
\resizebox{\linewidth}{!}{%
\begin{tabular}{lccccc}
\toprule
Scene & Proxy target & Aes3DGSNet & GSOQA & \shortstack{Stochastic\\NR-PCQA} & LRL-GQA \\
\midrule
\rowcolor{blockB}
A: low & .250 & .283 & .327 & .322 & .330 \\
\rowcolor{blockC}
B: medium & .422 & .457 & .438 & .464 & .467 \\
\rowcolor{blockD}
C: high & .638 & .555 & .468 & .486 & .457 \\
\bottomrule
\end{tabular}%
}
\end{table}

All methods separate scene A from scenes B and C. Aes3DGSNet has the widest prediction range across these examples (.272), closer to the target range (.388). The other methods give more compressed scores, especially for scene C. Our score for C is still below its proxy target (.555 vs. .638), showing a remaining gap at the high end. LRL-GQA also reverses the order of scenes B and C.

\end{document}